\documentclass[lettersize,journal]{IEEEtran}

\usepackage{tikz}
\usetikzlibrary{mindmap,trees,shapes.geometric, arrows}

\tikzstyle{startstop} = [rectangle, rounded corners, 
minimum width=3cm, 
minimum height=1cm,
text centered, 
text width=3cm,
text centered, 
draw=black, 
fill=red!30]

\tikzstyle{io} = [trapezium, 
trapezium stretches=true, 
trapezium left angle=70, 
trapezium right angle=110, 
minimum width=3cm, 
minimum height=1cm,text width=3cm, text centered, 
draw=black, fill=blue!30]

\tikzstyle{process} = [rectangle, 
minimum width=3cm, 
minimum height=1cm, 
text centered, 
text width=3cm, 
draw=black, 
fill=orange!30]

\tikzstyle{decision} = [diamond, 
minimum width=3cm, 
minimum height=1cm, 
text centered, 
draw=black, 
fill=green!30]
\tikzstyle{arrow} = [thick,->,>=stealth]

\usepackage{amsmath,amsfonts}
\usepackage{algorithmic}
\usepackage{algorithm}
\usepackage{array}
\usepackage{subcaption}
\usepackage{textcomp}
\usepackage{stfloats}
\usepackage{url}
\usepackage{verbatim}
\usepackage{graphicx}
\usepackage{cite}
\usepackage{forest}
\usepackage{hyperref}
\usepackage[capitalize]{cleveref}
\crefname{section}{Sec.}{Secs.}
\Crefname{section}{Section}{Sections}
\Crefname{table}{Table}{Tables}
\crefname{table}{Tab.}{Tabs.}
\usepackage{xcolor}
\usepackage{amssymb}
\usepackage{epstopdf}
\usepackage{multirow}
\usepackage{lscape}
\usepackage{graphicx}
\usepackage{rotating}
\usepackage{adjustbox}
\usepackage{booktabs}
 \usepackage{soul}
\usepackage[most]{tcolorbox}
\usepackage[numbers]{natbib}
\usepackage{csquotes}

\newtcolorbox{highlighted}{colback=yellow,coltext=red,breakable}
\usepackage{lipsum}
\newcommand{\HL}[1]{#1}
\DeclareMathOperator{\sign}{sign}
\begin{document}

\title{Vision-based Multi-future Trajectory Prediction: A Survey}
\author{Renhao Huang, Hao Xue, Maurice Pagnucco, ~\IEEEmembership{Member, IEEE}, Flora Salim, ~\IEEEmembership{Member, IEEE}, Yang Song
\thanks{The authors are with the School of Computer Science and Engineering, University of New South Wales, Sydney, Australia (Email:\{renhao.huang, hao.xue, morri, flora.salim, song.yang1\}@unsw.edu.au)}
}

\markboth{IEEE TRANSACTIONS ON NEURAL NETWORKS AND LEARNING SYSTEMS}%
{Shell \MakeLowercase{\textit{et al.}}: A Sample Article Using IEEEtran.cls for IEEE Journals}


\maketitle

\begin{abstract}
Vision-based trajectory prediction is an important task that supports safe and intelligent behaviours in autonomous systems. Many advanced approaches have been proposed over the years with improved spatial and temporal feature extraction. However, human behaviour is naturally diverse and uncertain. Given the past trajectory and surrounding environment information, an agent can have multiple plausible trajectories in the future. To tackle this problem, an essential task named \textit{multi-future trajectory prediction} (MTP) has recently been studied. This task aims to generate a diverse, acceptable and explainable distribution of future predictions for each agent. In this paper, we present the first survey for MTP with our unique taxonomies and a comprehensive analysis of frameworks, datasets and evaluation metrics. We also compare models on existing MTP datasets and conduct experiments on the ForkingPath dataset. Finally, we discuss multiple future directions that can help researchers develop novel multi-future trajectory prediction systems and other diverse learning tasks similar to MTP.
\end{abstract}


\section{Introduction}

\IEEEPARstart{V}{ision-based} trajectory prediction aims to predict the future trajectories of road users such as vehicles, pedestrians and cyclists based on their historical movement and their surrounding environments including terrain, obstacles and surrounding moving agents. The trajectories and scenes are often obtained from videos, GPS locations, or lidars. It is an essential task in autonomous driving, service robotics, crowd management and surveillance systems \cite{survey-htp}. 

Traditional methods explore physical models to model human behaviours. For example, Social Force \cite{social-force} uses attractive and repulsive forces to model social behaviours such as clustering and collision avoidance. However, such models have difficulties handling complex interactions to derive human-like future predictions. Recent data-driven methods learn complex spatial and temporal interactions from datasets. Many deep learning-based methods propose advanced modules such as \textit{pooling} \cite{social-lstm}, \textit{attention} \cite{social-gan} and \textit{graph neural networks} \cite{stgcnn, stgat} for social interaction. Meanwhile, to integrate scene features, some methods perform \textit{agent-centric alignment} \cite{mggan-tp} on RGB images or construct trajectory heatmaps \cite{ynet} or grid-occupancy maps \cite{scene-lstm}. For HD maps, some methods perform rasterisation \cite{tp-raster} with convolutions or \textit{polyline encoders} \cite{vectornet, lanegcn}.
\begin{figure}
    \centering
    \includegraphics[width=0.9\columnwidth]{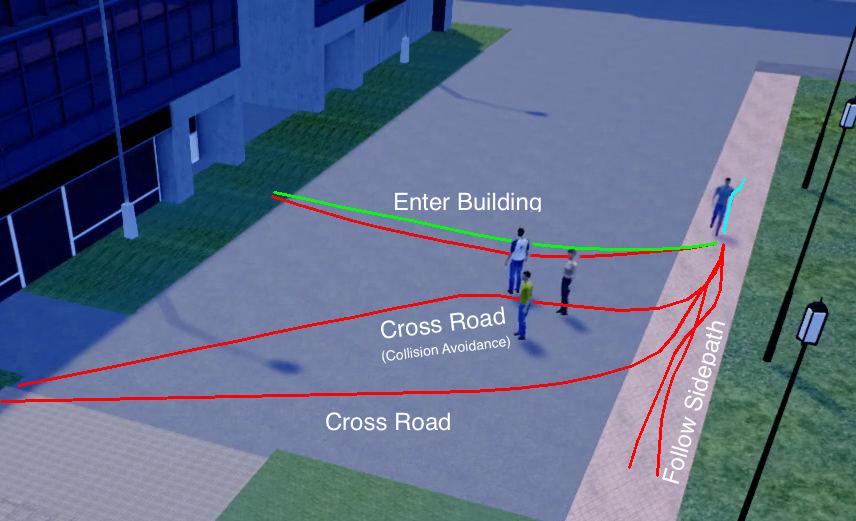}
    
    \caption{\textbf{An example of Multi-future Trajectory Prediction}. The blue, green and red lines are the observed, ground truth and other possible paths respectively. The agent can have multiple plausible trajectories including the ground truth given an observed path. }
    \label{fig:overview}
\end{figure}

\subsection{Uncertainty of Human's Future Behaviour} 
Human future behaviour inherently carries uncertainty that cannot be perfectly predicted. Given the observed information such as the past movement and surrounding information, \textit{there can be multiple plausible and socially acceptable behaviours in the future.} \cref{fig:overview} provides an illustration of this concept, where the target pedestrian can either enter the buildings, cross the road or continue to follow the side path, and all of these future trajectories are plausible given the observed trajectories and scene information. Most of the existing studies in pedestrian and vehicle trajectory prediction tasks \cite{desire, social-gan, mid} also use \textit{multimodality} to describe such uncertainty.

 
\subsection{Multi-future Trajectory Prediction}

 Considering that multiple future solutions are plausible, it is unrealistic for a model to accurately predict a single trajectory accurately without knowing humans' real intentions. Therefore, \textit{it is far better to foresee even without certainty than not} \cite{desire}. This leads to the introduction of the task known as a task named \textit{multi-future trajectory prediction} (MTP), also quoted as \textit{multimodal/diverse/stochastic trajectory prediction} in \cite{social-gan, desire, mid}. This task establishes a prediction protocol where multiple future trajectories are required to be predicted for each agent to reflect the distribution of future movement. Conversely, we use \textit{single-future trajectory prediction} (STP) to describe trajectory prediction tasks that generate only one proposed future trajectory. As a more realistic problem in trajectory prediction, MTP has become the default setting in almost all recent studies \cite{stgat, densetnt, vectornet, qcnet}. MTP is expected to meet the following goals (see \cref{fig:mtp_expect} for examples):
\begin{itemize}
	\item \textit{Accuracy}, where the ground truth should be covered by the predicted distribution. This is a basic goal in all prediction problems and MTP model should fully use the observed information for predictions.

	\item \textit{Diversity}, where the predicted distribution is expected to cover all possible solutions. In other words, MTP models should avoid the mode collapse problem and predictions should cover all possible paths.
		
	\item \textit{Social Acceptance}, where the predicted paths should be realistic with the past trajectory and follow social norms, e.g., collision-free predictions and motion changes in different terrains. This requires the MTP model to correctly learn the interactions between agents and surrounding environments.
		
	\item \textit{Explainability/Controllability}, where predictions should be controllable and understandable to humans. This requires the MTP models to provide explanations for the predicted prediction. 
				
\end{itemize}
Furthermore,  all model optimisation and evaluation can use only one ground truth trajectory. Numerous methods have been developed to address this problem and better estimate the distribution. Apart from these, advanced datasets and evaluation metrics have been proposed to enhance the evaluation of MTP.

%
 
\subsection{A Survey for MTP}
 
This paper presents the first survey for MTP, which delves into papers that propose distinctive frameworks, datasets and evaluation metrics from the \textit{multi-future prediction} perspective. Our investigation specifically categorises current MTP methods into \textit{Noise-based}, \textit{Anchor-conditioned}, \textit{Recurrent-based} frameworks, alongside other special improvement techniques, offering a comprehensive analysis of each category. Subsequently, we present the datasets often used for the evaluation of MTP. Moreover, we present evaluation metrics specialised for MTP, including \textit{lower-bound-based}, \textit{probability-aware} and \textit{distribution-aware} metrics. Finally, we summarise multiple novel directions for future MTP works. In summary, our contributions can be summarised into following aspects:
	\begin{enumerate}
		\item To our knowledge, this is the first survey of trajectory prediction to categorise current methods from the multi-future trajectory prediction perspective.
		\item We review and provide our unique taxonomies for the deep learning frameworks, datasets and evaluation metrics for MTP and analyse their advantages and issues.
            \item We compare typical MTP models on commonly used benchmarks and report distribution-aware MTP metrics on the ForkingPath dataset.
		\item We discuss potential directions for MTP that should be the focus of future research and other diverse learning tasks that can be inspired from MTP.
	\end{enumerate}
\begin{figure*}[t]
	\centering
	\includegraphics[width=0.90\textwidth]{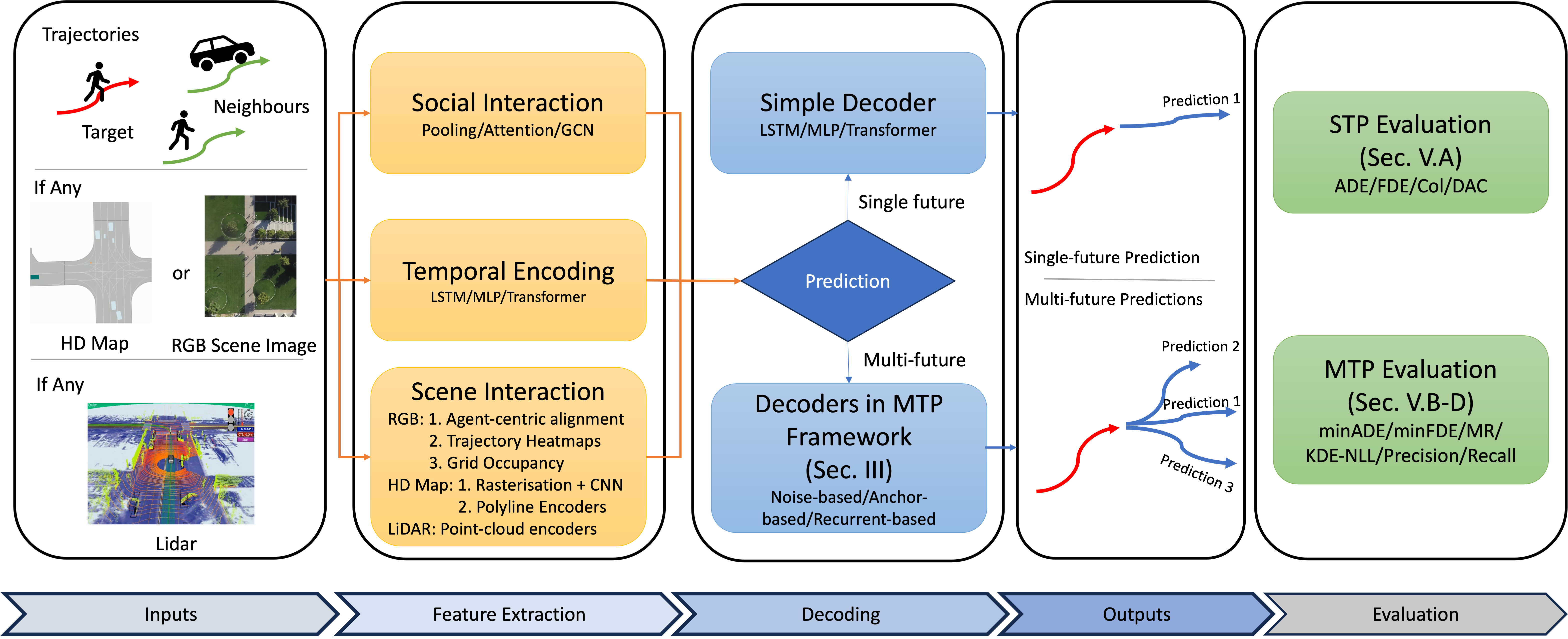}
	\caption{Background: pipeline of deep learning-based trajectory prediction and the role of MTP. This survey mainly focuses on the decoders in MTP frameworks and MTP Evaluation.}
	\label{fig:pipeline-tp}
\end{figure*}

\noindent \textbf{Comparison with existing surveys.} 
Existing surveys have conducted comprehensive investigations from physical models to deep learning models for pedestrian \cite{survey-htp,jiang2025survey} and vehicle \cite{survey-vtp, review-vbp} trajectory prediction. Some surveys adopt diverse perspectives such as risk assessment \cite{survey-mp-risk}, social-interaction \cite{survey-social-interaction}, knowledge incorporation \cite{survey-vtp-knowledge}, intersection monitoring \cite{survey-intersection-monitoring} and different modalities of inputs \cite{survey-multimodal-data}. A recent survey for autonomous driving \cite{survey-av} summarises trajectory prediction frameworks from physical models to deep learning approaches. In addition, Wang \textit{et al.} \cite{survey-multimodal-vehicle-tp} propose a survey investigating and comparing vehicle trajectory prediction models with interaction modelling techniques. However, these survey papers \textbf{all} build their taxonomies from a feature extraction perspective, some only containing a brief description of MTP as auxiliary content. This paper is the first survey specifically focusing on the trajectory prediction methods that propose special frameworks, datasets and evaluation metrics for MTP. 

This survey is organised as follows. \Cref{sec:general-framework} introduces the general framework for trajectory prediction including the problem definition and pipeline of deep-learning trajectory prediction and illustrates the role of MTP in it. \Cref{sec:methods}, \Cref{sec:datasets} and \Cref{sec:evaluation-metrics} present the frameworks, datasets \& benchmarks and evaluation metrics for MTP. Finally, \Cref{sec:discussion} discusses the future work for MTP.
 \section{Background: General Framework of Deep Learning-based Trajectory Prediction}
\label{sec:general-framework}
Deep learning-based trajectory prediction models mostly follow a sequence-to-sequence framework as shown in \cref{fig:pipeline-tp}. To build a trajectory prediction model, three types of feature extraction modules are needed for (1) social interaction modelling, by understanding how neighbouring agents can affect the movement of target agents; (2) temporal encoding, which aims to obtain the information from target agents' past trajectories; and, (3) scene interaction modelling, which explores how surrounding environments (terrains, buildings) can affect target agents' future routes. Subsequently, the extracted features are passed into either the STP or MTP decoders. These decoders are designed to generate single or multiple trajectories respectively and the resultant predictions are finally evaluated by corresponding metrics.

\subsection{Input Data}

\noindent \textbf{Agent.} In trajectory prediction tasks, we use \textit{agents} to describe road users with self-cognition, which can be pedestrians, drivers, motorists or cyclists. We also use \textit{target agents} to denote the agents of interest for prediction and \textit{neighbouring agents} to denote other road users.

\noindent \textbf{Trajectory.} A \textit{trajectory} of an agent $i$ in trajectory prediction is defined as a sequence of real-world or pixel coordinates: $\{X_i^T, Y_{i}^\tau\}$, where \[X_i^T = \{X_i^t | t\in[1,2, \cdots, T_{obs}]\}\] is the observed trajectory with  $T_{obs}$ timesteps and \[Y_{i}^\tau=\{Y_{i}^t | t\in (T_{obs},T_{obs}+1,\cdots, T_{obs}+T_{pred}]\}\] is the ground truth of the future path with $T_{pred}$ time steps. Here $i$ is the index among $N$ agents in a scene $i \in [1,2,\cdots,N]$. Both $X_i^t$ and $Y_i^t$ are conventionally 2D coordinates in most trajectory prediction tasks but can be extended to multi-dimensional coordinate spaces. For models requiring social interaction handling, neighbouring agents' observed trajectories are also provided as $X_{1:N\setminus i}^T$, denoting a set of observed trajectories within a slot of frames excluding the target agents. 

\noindent \textbf{Coordinate System.} Trajectories obtained from the object tracking on LiDAR data \cite{waymo, argoverse, nuscenes, jrdb} are in a \textit{real-world coordinate system}, indicating real-world distances in meters. Trajectories obtained from videos \cite{sdd, eth, ucy, virat} are initially in a \textit{pixel coordinate system} but can be converted to real-world coordinates using the camera parameters. Alternatively, real-world distance can be estimated based on the ratio of objects \cite{eth,ucy}. For studies on the first-person videos \cite{jaad,pie, titan}, they predict trajectories directly on pixel coordinates.

\noindent \textbf{Scene Information.} Some models need to generate future paths compliant with the geometric terrain. In pedestrian trajectory prediction, images contain information about the static environment such as the building, pavement and parking slots. Some works further convert these images to segmentation maps to simplify the semantic information. In autonomous driving datasets, high-definition (HD) maps are provided to represent topological information of the road such as the lane and traffic lights. Some models directly use these raw data \cite{tnt,vectornet} while others convert them to a rasterised representation \cite{multipath}. LiDAR data are also useful information in autonomous driving \cite{waymo, argoverse,nuscenes} or social robots \cite{jrdb, sit}, which are usually presented as point clouds. These data construct 3D geometries of the surrounding environment and provide more accurate and comprehensive information than scene images and HD maps.

\subsection{Overall Problem Definition} 
\label{sec:problem}
The goal of \textit{trajectory prediction} is to optimise a model $f_{TP}$ to predict $K$ future trajectories: \[\hat{Y}_{i,K}^\tau = \{\hat{Y}_{i,k}^t|k=1,2, \cdots, K\}\] using observed information $X_i^T, X_{1:N\setminus i}^T, S$ and a controlling variable $c_K$ as inputs: 

\begin{equation}
	\hat{Y}_{i, K}^\tau = f_{TP}(X_i^T, X_{1:N\setminus i}^T, S, c_K)
	\label{eq:problem-definition}
\end{equation}
where $X_{1:N\setminus i}^T$ are the set of agents $i$'s neighbours' observed trajectories and $S$ is the scene information. When $K=1$, only a single prediction is allowed for each agent. We also use $c_K$ to denote a set of conditional variables that control the predictions. When it is ignored ($c_K=\varnothing $), the task is \textit{single-future trajectory prediction (STP)} and expects a minimum prediction error compared to $Y_i^\tau$. Otherwise, it becomes \textit{multi-future trajectory prediction (MTP)} and aims to predict a distribution of all acceptable future trajectories, each controlled with a unique condition $c_k \in c_K$.

\subsection{Feature Extraction}
In trajectory prediction, there are many studies proposing advanced feature extraction modules to obtain representative information from observed data. Techniques for feature extraction can be commonly applied in both MTP and STP.

\noindent \textbf{Temporal Encoding.} This part aims to extract temporal features from agents' observed trajectories. Most works use recurrent neural networks such as Long Short-term Memory (LSTM) \cite{social-lstm, ss-lstm, scene-lstm, social-gan}, Gated Recurrent Unit (GRU) \cite{desire} and Transformer encoder \cite{tf-traj, star} to extract temporal information from sequential coordinates. Some works directly use Multi-layer Perceptron (MLP) \cite{pecnet} and Temporal CNN \cite{cnn-traj, vectornet} to simplify the training. Meanwhile, some methods convert coordinates into heatmaps and use 2D convolutions to extract temporal features.

\noindent \textbf{Social Interaction.} When agents walk or drive in open places, it is necessary to consider other agents and adjust the routes. This includes collision avoidance and group/leader following or merging \cite{trajnet++}. To model such interaction, given the target and neighbouring agents' temporal features, some studies propose pooling-based \cite{social-lstm, social-gan}  and attention-based mechanisms \cite{social-attention, sophie}. Some methods \cite{stgat, social-bigat, stgcnn} further construct graphs to describe the social interaction where vertices are the temporal features and use graph neural networks to pass messages among nodes and extract node features. 

\noindent \textbf{Scene Interaction.} Pedestrians typically walk in side paths and drivers should drive in lanes. Therefore, agents need to consider the terrain semantics to decide the routes, described as scene interaction. On image data, most methods \cite{sophie,carnet,ynet,ss-lstm,mggan-tp, goal-gan} use CNNs to extract scene features from images. To fuse the scene and trajectory features, some methods \cite{sophie,mggan-tp, goal-gan} perform the agent-centric alignment which centralises, rotates and crops the images to agents' last observed position and heading direction and fuse them with other features via soft-attention while YNet \cite{ynet} transforms the 2D coordinates into heatmaps and concatenate them with the scene image. Some methods \cite{scene-lstm, safecritic} manually build grid occupancy for the moving agents to represent the walkable and non-walkable places. In vehicle trajectory prediction tasks, some methods convert HD maps to rasterisation maps and encode them with convolutions while some works \cite{vectornet, lanegcn, densetnt} directly use polyline encoders (e.g., GNN) to extract the lane information efficiently. For point-cloud data from LiDAR, point-cloud encoders are usually used \cite{jrdb,sit}.

\subsection{Decoding} Decoders for STP simply roll out the extracted features to generate a single future trajectory. Early decoders are based on LSTM to model the sequential outputs \cite{social-lstm, ss-lstm}. Advanced STP decoders use a Transformer-based decoder \cite{tf-traj}. Recently, some works \cite{nar-traj} explore advanced non-autoregressive decoders to predict future locations in parallel. Decoders for MTP can adapt one of the decoders for STP with MTP frameworks, which will be described in detail in \Cref{sec:methods}. 
	
\subsection{Relationship between STP and MTP}

Based on well-designed feature encoders and trajectory decoders,  MTP requires advanced frameworks to fully use these features to generate multiple diverse and high-quality predictions. As shown in \cref{eq:problem-definition}, MTP frameworks propose multiple predictions with additional controlling variables $c_K$ as inputs. In other words, frameworks for MTP explore advanced $c_K$ to better control the predicted future trajectories. For example, $c_K$ can be Gaussian noise in (\cref{sec:noise-mtp}), endpoints, prototype trajectories in (\cref{se:anchor-mtp}) and different previous states in (\cref{sec:rnn-mtp-framework}). In addition, different MTP frameworks can have advanced training strategies as well as different sampling tricks to better train the model.

 \begin{figure*}[t]
    \centering
    \resizebox{0.95\textwidth}{!}{
	\begin{forest}
  for tree={
      grow=east,
      reversed=true,
      anchor=base west,
      parent anchor=east,
      child anchor=west,
      base=left,
      font=\small,
      rectangle,
      draw,
      rounded corners,align=left,
      minimum width=2.5em,
      inner xsep=4pt,
      inner ysep=0.5pt,
  },
  where level=1{fill=blue!10,align=center}{},
  where level=2{font=\footnotesize,fill=pink!30,align=center}{},
  where level=3{font=\footnotesize,yshift=0.26pt,fill=yellow!20}{},
    [MTP \\ Frameworks, fill=gray!20
        [Noise-based Framework,
          [\emph{GAN} 
            [ \emph{GAN}: SGAN \cite{social-gan}; \emph{InfoGAN}: S-Ways \cite{socialways}; \\ \emph{BicycleGAN}: Social-BiGAT \cite{social-bigat}; \emph{MGAN}:MG-GAN \cite{mggan-tp};\\
          	\emph{Discriminators}:
                \cite{gan-for-human-traj-pred}; \cite{se-gan}; \cite{safecritic}; \cite{sgan-v2}
            ]
          ]
          [\emph{CVAE} 
            [Desire \cite{desire}; AgentFormer \cite{agentformer}; DisDis \cite{pcmd};\\ ABC \cite{action-contrastive-learning}; SocialVAE \cite{social-vae}]
          ]
          [\emph{NF}
            [HBAFlow \cite{hba-flow};  FloMo \cite{flomo}; STGlow \cite{stglow}]
          ]
          [\emph{DDPM} 
            [MID \cite{mid}; MotionDiffuser \cite{motion-diffuser}; LED \cite{led}; \\ SingularTrajectory \cite{singular-trajectory}; SPDiff\cite{chen2024social};]
          ]
        ]
        [Anchor Conditioned Framework
            [\emph{PEC}, font=\footnotesize
            	[PECNet \cite{pecnet};  DenseTNT \cite{densetnt}; TPNet \cite{tpnet}; 
                TNT \cite{tnt}; \\ YNet \cite{ynet}; Home \cite{home}; ExpertTraj \cite{experttraj}; \\HyperTraj \cite{hypertraj}; DecoupleTraj \cite{decouple-traj}; GANet \cite{ganet}; \HL{CGTP \cite{cgtp}}]
            ]
            [\emph{PTC}, font=\footnotesize
                [ 	CoverNet \cite{covernet}; MultiPath \cite{multipath}; mmTransformer \cite{mmtransformer}; \\
               		S-Anchor \cite{social-anchor}; SIT \cite{social-interpretable-tree}; PCCSNet \cite{pccsnet};\\
  					MultPath++ \cite{multipath-pp}; \quad QCNet \cite{qcnet}; ProphNet \cite{prophnet}; \\ EigenTrajectory \cite{eigentrajectory}; SICNet \cite{sicnet}; SingularTrajectory \cite{singular-trajectory}
                ]
            ] 
        ]
        [Recurrent-based Framework
            [\emph{Gaussian Sampling} [Trajectron \cite{trajectron}; Trajectron++ \cite{trajectron-pp}], fill=yellow!20]
            [\emph{Grid Sampling} [Multiverse \cite{multiverse}; ST-MR \cite{stmr}; TDOR \cite{tdor}], fill=yellow!20]
            [\emph{Token Sampling} [LMTraj \cite{lmtrajectory}], fill=yellow!20]
        ]
        [Other Techniques \\for Improved MTP
            [Variety Loss
        		[Analysis: \cite{variety-loss-analysis}; Usage:\cite{social-gan,tp-raster, ewta,lanegcn}.]
        	]
        	[Bivariate Gaussian Outputs
        		[STGCNN \cite{stgcnn}; SGCN\cite{sgcn}; Trajectron \cite{trajectron}; \\ Trajectron++ \cite{trajectron-pp}; MultiPath \cite{multipath}]
        	]
            [Diverse Sampling Tricks
                [TTST \cite{ynet}; LDS \cite{likelihood-diverse-sampling}; Generative Style Decoder (GSD) \cite{tdor};\\ NPSN \cite{npsn}; Stimulates Verification (SV) \cite{stimulus-verification}; \\ BOSampler \cite{bosampler}]
            ]
            [Joint Sampling Tricks
                [Fictitious Play \cite{tp-fictitious-play}; M2I \cite{m2i}; SceTP \cite{scept}; \\CGTP \cite{cgtp}; IPCC-TP \cite{ipcc-tp}]
            ]
        ]
    ]
\end{forest}

    }
    \caption{An overview of the taxonomy of MTP frameworks.}
    \label{fig:mindmap-frameworks}
\end{figure*}

\begin{figure*}[t]
    \centering
     \begin{subfigure}[b]{0.24\textwidth}
         \centering
         \includegraphics[height=\textwidth]{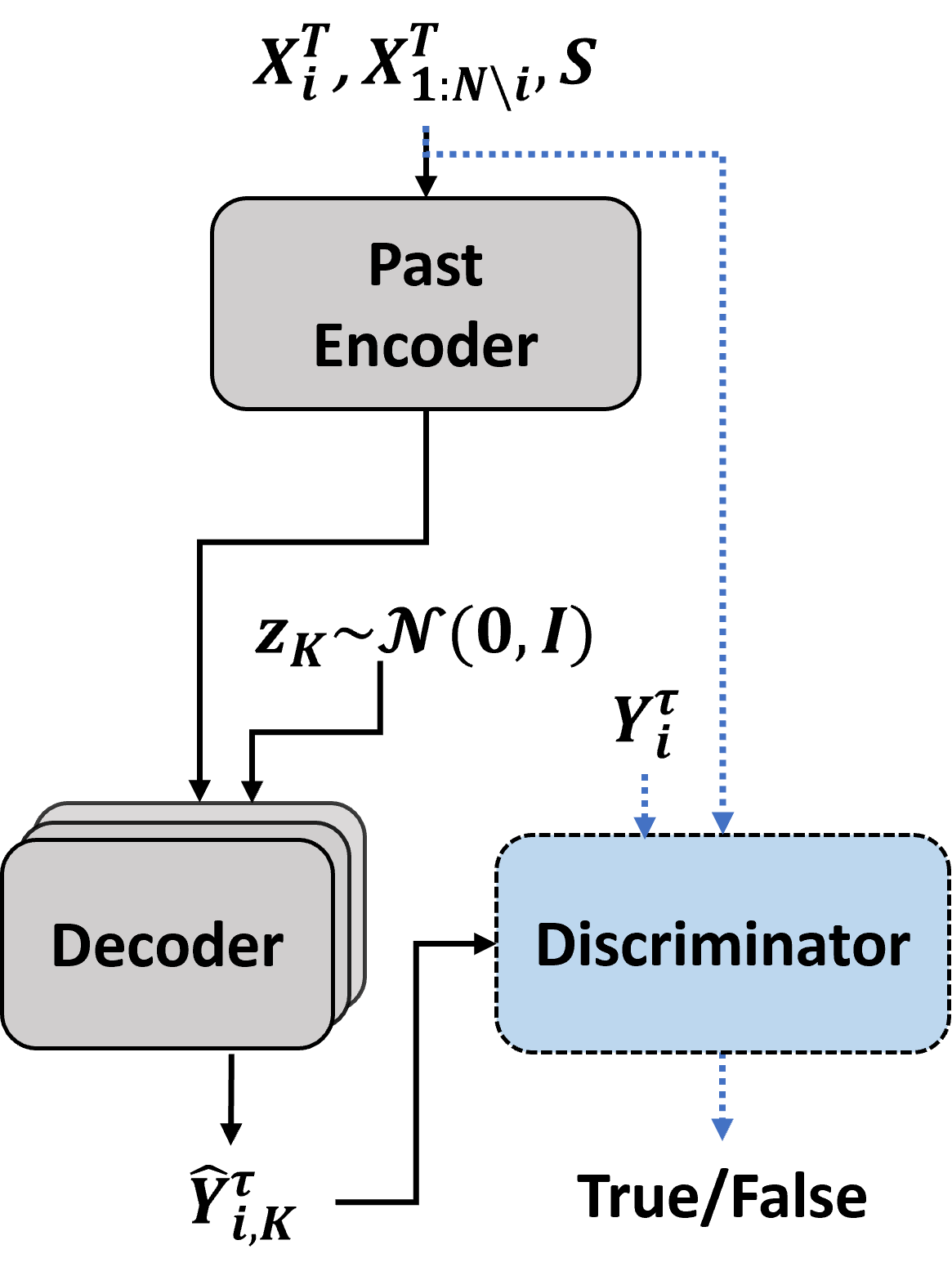}
         \caption{Noise-based: GAN}
         \label{fig:gan}
     \end{subfigure}
     \begin{subfigure}[b]{0.24\textwidth}
         \centering
         \includegraphics[height=\textwidth]{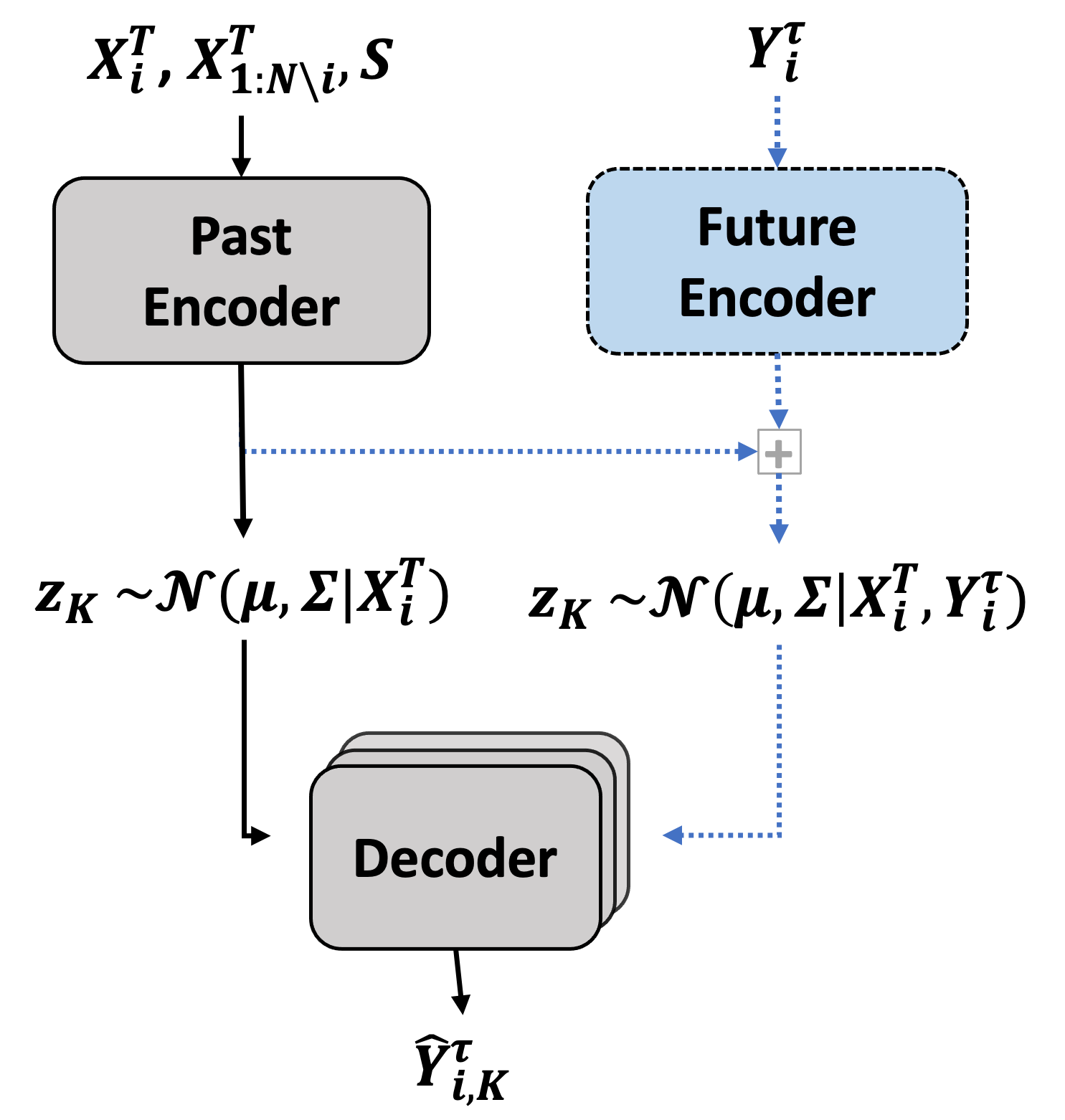}
         \caption{Noise-based: CVAE} 
         \label{fig:cvae}
     \end{subfigure}
      \begin{subfigure}[b]{0.24\textwidth}
         \centering
         \includegraphics[height=\textwidth]{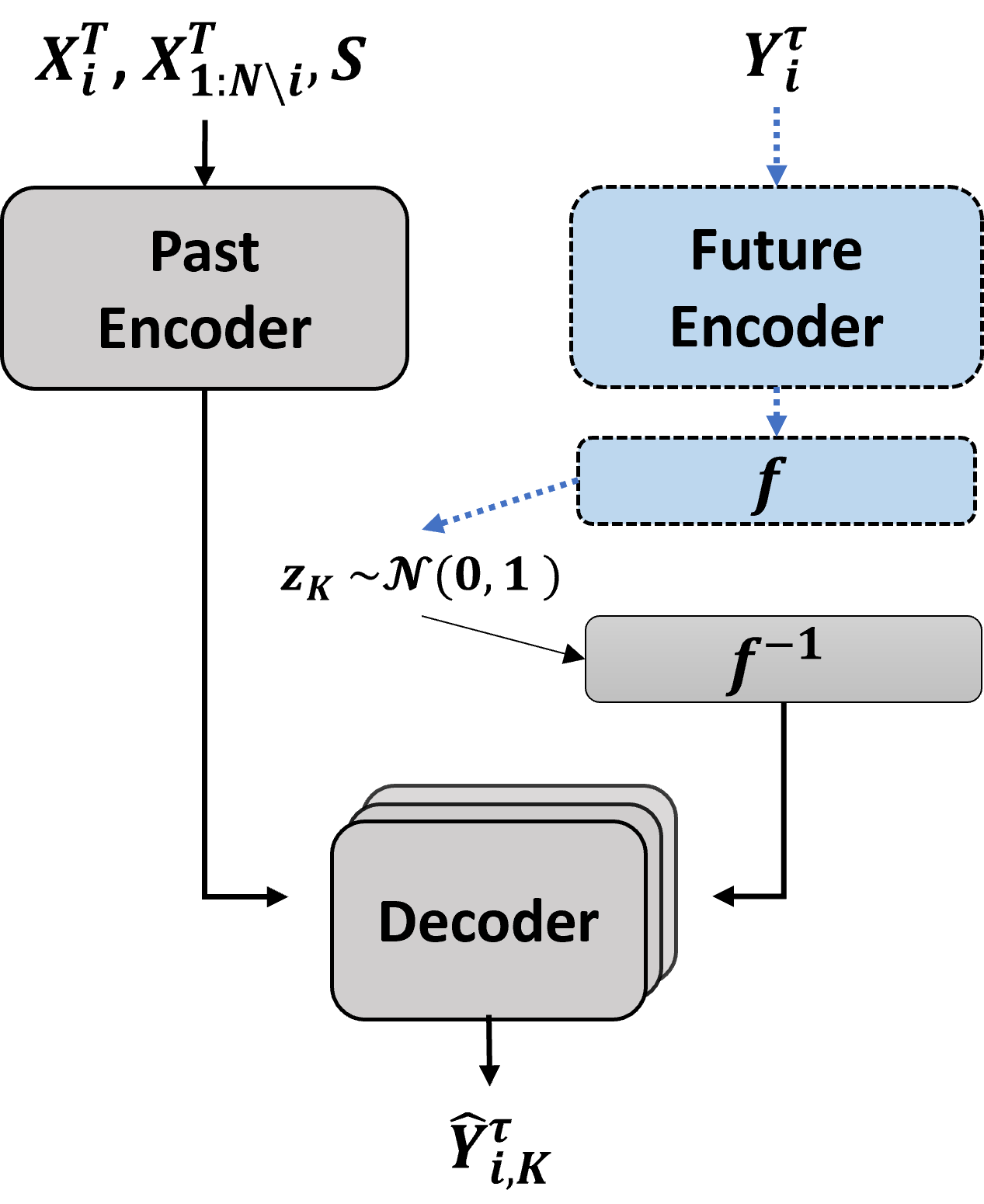}
         \caption{Noise-based: NF}
         \label{fig:nf}
     \end{subfigure}
     \begin{subfigure}[b]{0.24\textwidth}
         \centering
         \includegraphics[height=\textwidth]{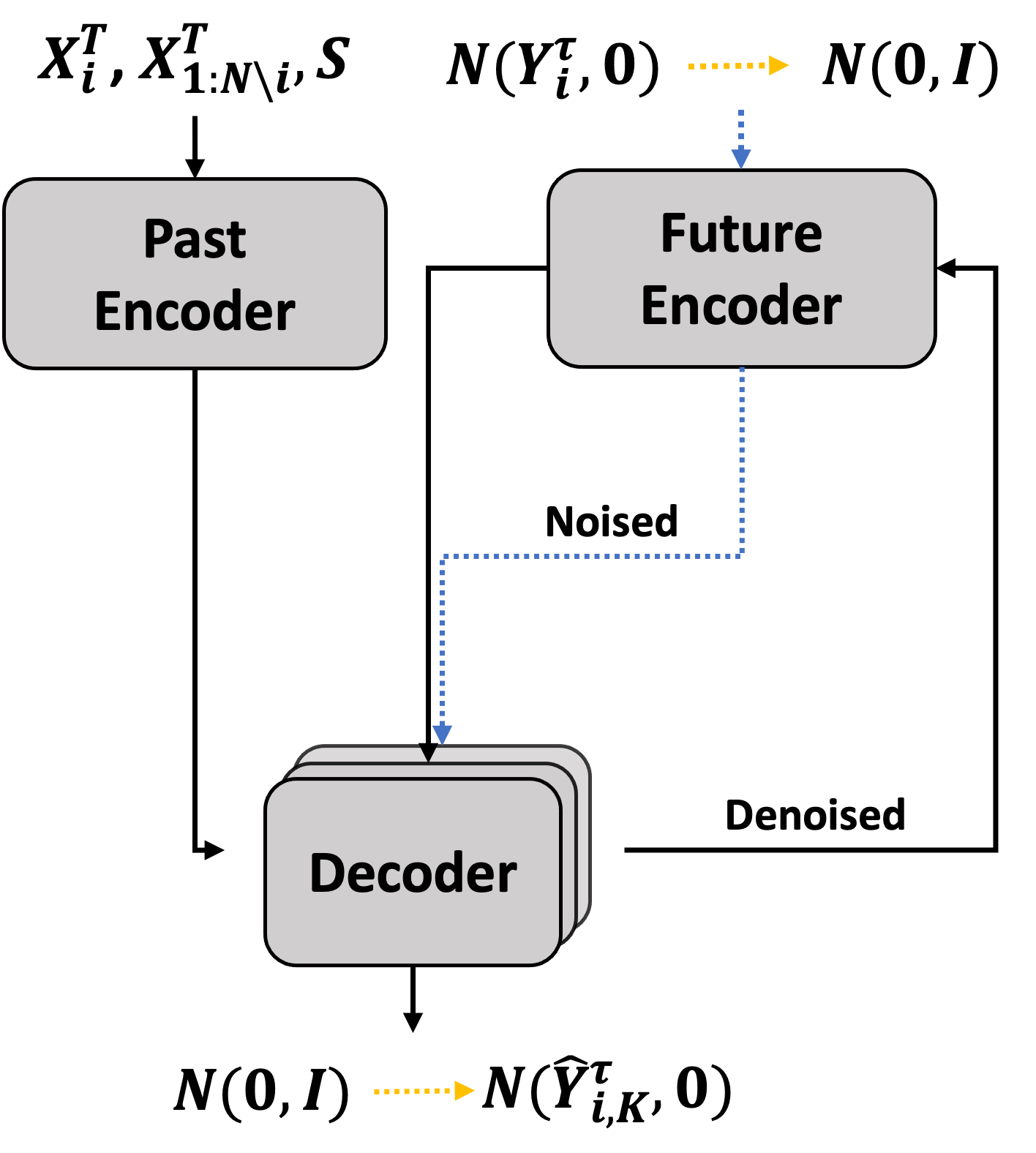}
         \caption{Noise-based: DDPM}
         \label{fig:nf}
     \end{subfigure}
     
     \begin{subfigure}[b]{0.3\textwidth}
         \centering
         \includegraphics[height=0.8\textwidth]{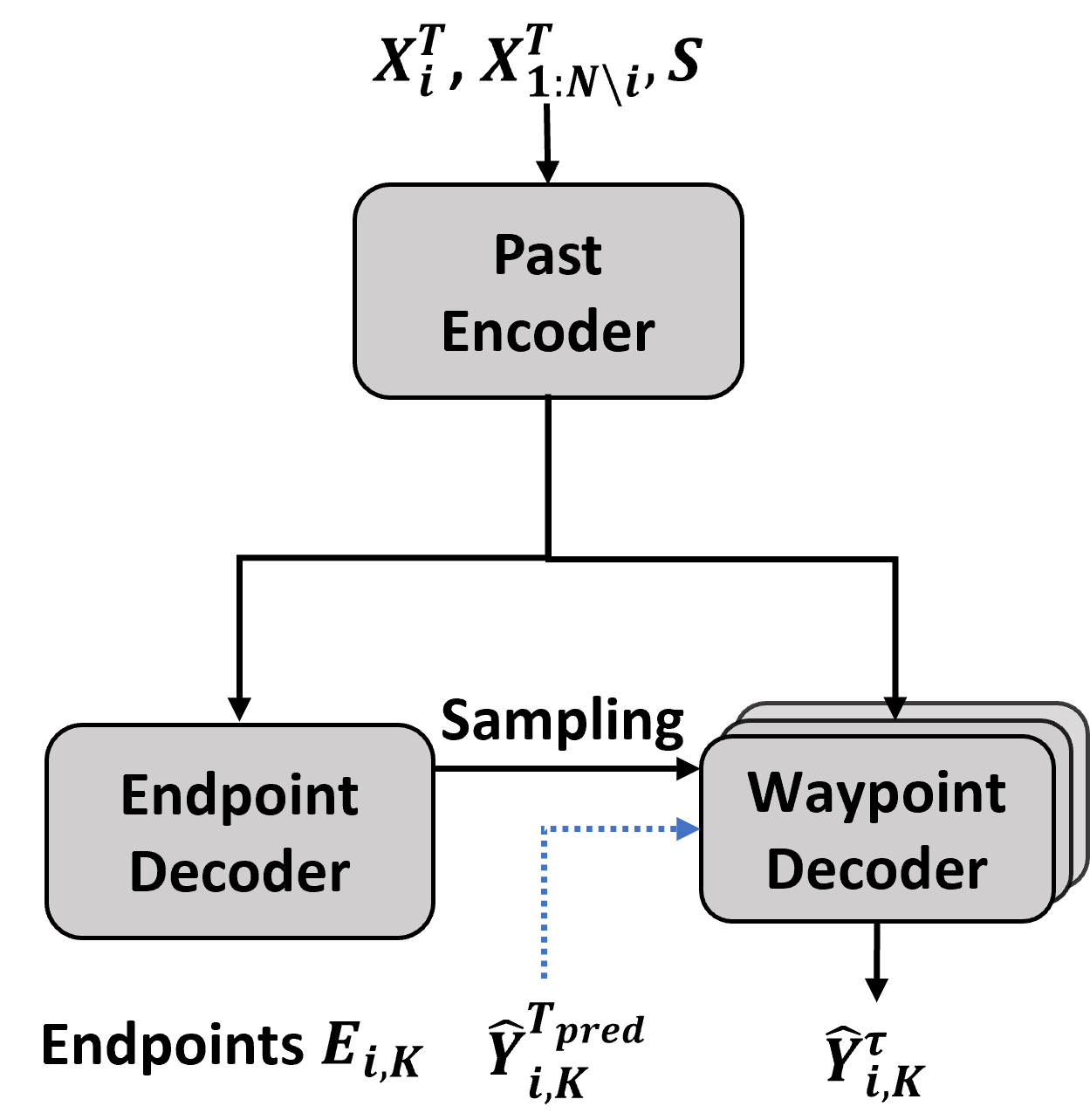}
         \caption{Anchor-conditioned: PEC}
         \label{fig:pec}
     \end{subfigure}
     \begin{subfigure}[b]{0.3\textwidth}
         \centering
         \includegraphics[height=0.8\textwidth]{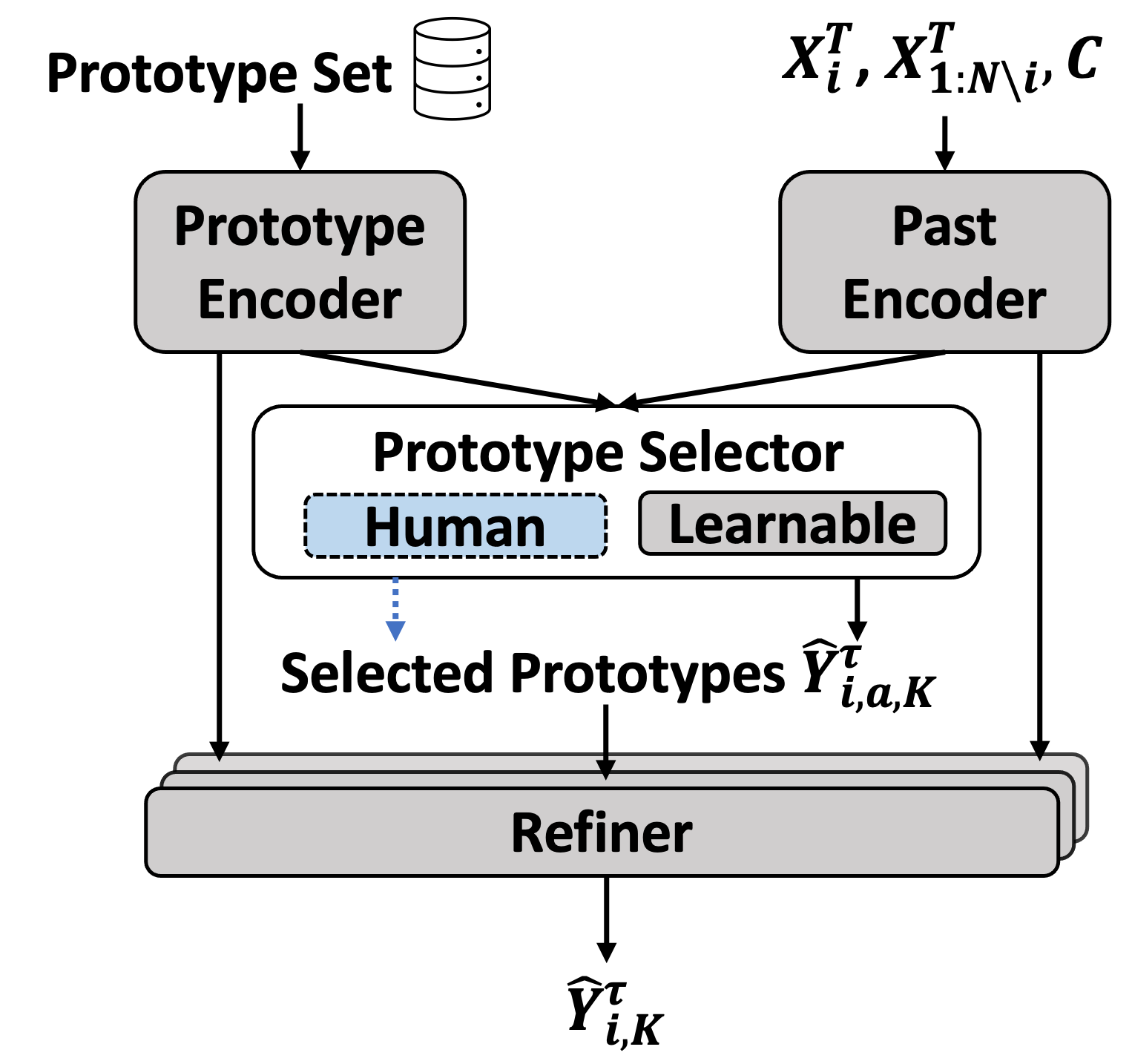}
         \caption{Anchor-conditioned: PTC}
         \label{fig:ptc}
     \end{subfigure}
     \begin{subfigure}[b]{0.30\textwidth}
         \centering
         \includegraphics[height=0.7\textwidth]{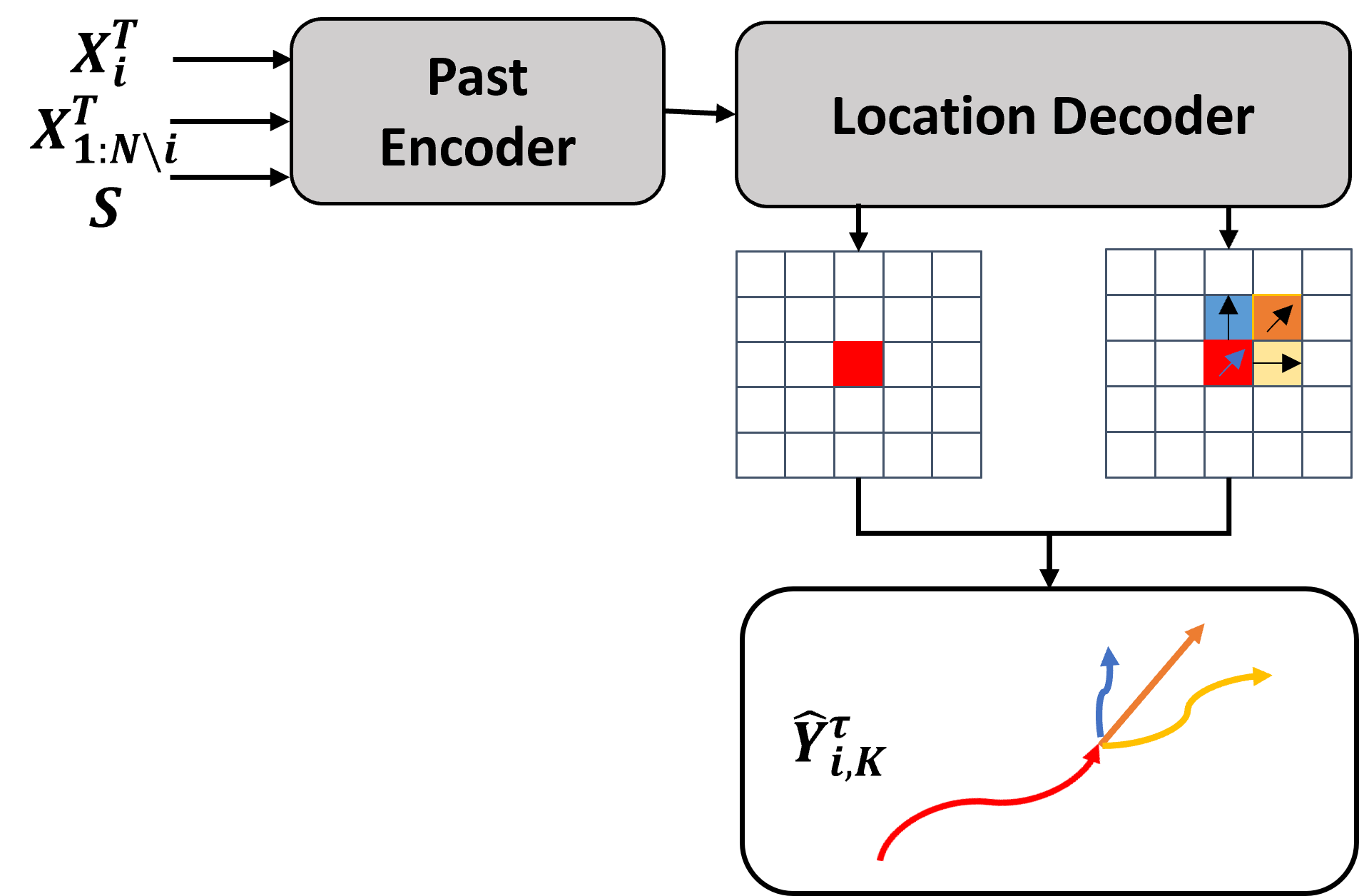}
         \caption{Recurrent-based: Grid Sampling}
         \label{fig:grid}
     \end{subfigure}
    
    \caption{ General Pipelines of MTP frameworks. The block with multiple layers is executed repeatedly with $K$ different inputs. Dashed blue lines and blocks are executed for training only.}
    \label{fig:frameworks}
\end{figure*}
\section{Frameworks for MTP}
\label{sec:methods}

In this section, we review the MTP frameworks with their taxonomy in \cref{fig:mindmap-frameworks} and general pipelines in \cref{fig:frameworks}.

\subsection{Noise-based MTP Framework}
\label{sec:noise-mtp}
The simplest way to convert STP to MTP is to inject random noise into the model. In this section, we discuss the \textit{Noise-based MTP framework} introduced by \cite{social-gan} where features from the past encoder are concatenated with Gaussian noise vectors and sent together into the decoder for diverse predictions. However, it may generate unrealistic predictions and is difficult to control. To tackle this problem, many advanced generative
frameworks are used to improve the model.

\noindent \textbf{Generative Adversarial Network (GAN).} 
The reconstruction loss aims to align the predictions with the ground truth. Nevertheless, it is challenging to describe ``good'' and ``bad'' predictions abstract to humans. One solution is to follow the adversarial training proposed in \cite{gan}, which involves training a discriminator to distinguish whether the given samples are from the dataset (real) or generated (fake) by the model. This discriminator evolves along with the generator and guides it to generate samples belonging to the distribution of the dataset. 

Following this strategy, Social-GAN \cite{social-gan} introduces the first \textit{GAN-based MTP framework} and the general pipeline is shown in \cref{fig:gan}. The trajectory prediction model $f_{TP}$ is learned collaboratively with a discriminator $D$ and the overall loss $L_{adv}$:
	\begin{equation}
	   \begin{split}
		L_{adv}(X_i^T, \hat{Y}_{K,i}^\tau,Y_i^\tau) = & \mathbb{E}_{Y_i^\tau, X_i^T}[\log D(Y_i^\tau, X_i^T)] + \\
				  & \mathbb{E}_{\hat{Y}_{k,i}^\tau,  X_i^T}[\log(1-D(\hat{Y}_{k,i}^\tau,X_i^T)]
            \end{split}	
        \end{equation}

Traditional GAN suffers from mode collapse due to the overfitting and the model tends to generate samples with a single mode. To solve this problem, further works use more advanced GAN frameworks. For example, Social-BiGAT \cite{social-bigat} follows the BicycleGAN \cite{bicycle-gan} to perform reversible transformations between each generation and its latent noise vector to further alleviate the mode collapse. Concurrently, Amirian \textit{et al.} \cite{socialways} claim that the reconstruction loss is the main factor that causes mode collapse. Hence, their model SocialWays depreciates this loss and follows InfoGAN \cite{infogan} to use latent code to control the prediction. Patrick \textit{et al.} \cite{mggan-tp} suggest that the manifold of the distribution of future paths is discontinued and thus cannot be well covered by other GAN-based methods. Therefore, they propose MG-GAN by using multiple decoders, each one handling a continuous sub-manifold. 

In addition, more advanced discriminators in GAN-based models have been proposed to improve the quality of generated trajectories. For example, \cite{gan-for-human-traj-pred} and \cite{se-gan} propose improved discriminators to simplify the adversarial training for recurrent neural networks. SC-GAN \cite{sc-gan} enhances the discriminator to check scene compliance using differentiable rasterised maps and scene images while some methods propose an improved discriminator to ensure social acceptance \cite{sgan-v2,safecritic}.

\noindent \textbf{Conditional Variational Autoencoders (CVAE).} 
CVAE-based trajectory prediction models \cite{desire,agentformer,pcmd} follow \cite{cvae-3}
that maximises the evidence lower bound of feature distribution as shown in \cref{fig:cvae} and is an alternative to encourage diverse predictions. Given the past encoder $p_{\theta}$, future encoder $q_{\Phi}$ and the decoder $p_{\phi}$, the loss function is as follows:
 \begin{equation}\label{eq1}
 \begin{split}
     L_{VAE} = & -D_{KL}(q_{\Phi}(z|X_i^T, Y_i^T) || p_{\theta}(z|X_i^T))  \\ 
     & + \mathbb{E}_{p_{\theta}(z|X_i^T)}[\log p_{\phi}(\hat{Y}^\tau_i|z)]
 \end{split}
 \end{equation}
where $D_{KL}$ is the Kullback-Liebler divergence.

In addition, the latent distribution of CVAE can be better controlled and enhanced. For instance, DisDis \cite{pcmd} and ABC \cite{action-contrastive-learning} predict personalised and action-aware motion patterns by distinguishing the feature distributions via contrastive learning. A recent model named SocialVAE \cite{social-vae}  uses a timewise CVAE by applying CVAE with recurrent networks. We strongly recommend \cite{cvae-traj} for a comprehensive review of CVAE in trajectory prediction.

\noindent \textbf{Normalising Flow (NF).} GAN or CVAE-based models are difficult to train due to their implicit distribution modelling. Therefore, the NF-based MTP framework is proposed to explicitly learn the data distribution through an invertible network shown in \cref{fig:nf}, converting a complicated distribution into a tractable form via invertible transformations. For example, HBAFlow \cite{hba-flow} uses a Haar wavelets-based block autoregressive model that splits couplings to learn distributions for motion prediction while FloMo \cite{flomo} utilises monotonic rational-quadratic splines for expressive and fast inversion. STGlow \cite{stglow} proposes the generative flow with pattern normalisation to learn the motion behaviour conditioned on the representation of social interactions. However, no NF-based model is capable of handling discontinued manifolds. A plausible solution is to follow \cite{mggan-tp} to use multiple invertible decoders.

\noindent \textbf{Denoising Diffusion Probabilistic Models (DDPM).} Diffusion model is a strong generative model in computer vision \cite{ddpm}. In MTP, MID \cite{mid} is firstly proposed for predictions with controllable diversity. It follows DDPM \cite{ddpm} that the latent vector $z_K$ is sampled from Gaussian distributions with controlled stochasticity via parameterised Markov Chain. LED \cite{led} accelerates the diffusion process by using a leapfrog initialiser to skip the denoising steps. This method alleviates the significant time consumption during the reverse diffusion process due to the required number of steps. MotionDiffuser \cite{motion-diffuser} proposes the first multi-agent DDPM framework by using cross attention to fuse the neighbour information. SingularTrajectory \cite{singular-trajectory} proposes to progressively refine prototype trajectories to a realistic trajectory through a diffusion process. Future studies for DDPM-based MTP can develop advanced mechanisms to control the generation and explore more techniques for acceleration. SPDiff \cite{chen2024social} incorporates social interactions and proposes multi-frame rollout training for long-term prediction. 

\subsection{Anchor Conditioned MTP Framework} 
\label{se:anchor-mtp}
To effectively guide the model to predict trajectories with controlled behaviours, it has been proposed that each prediction can be conditioned on a prior \cite{multipath, tnt}, also named \textit{anchors}, which are explicit to each possible trajectory and provide large information gain to the predictions. Well-known anchors include the following:
	\begin{itemize}
		\item \textit{Endpoints} \cite{pecnet,ynet,tnt,densetnt}: the final locations the agent will arrive at; and
		\item \textit{Prototype trajectories} \cite{multipath, covernet, qcnet}: typical types of motion patterns explicit to agents' different intentions. 
	\end{itemize}

We categorise the methods using anchors as \textit{anchor conditioned MTP framework}. This framework usually contains two sub-tasks: (1) \textit{anchor selection}, which selects $K$ plausible anchors from an anchor set and; (2) \textit{waypoint decoding}, which predicts \textit{waypoints}, the final prediction of future trajectory, based on the given anchor. The anchor selection can be performed via random sampling or top $K$ ranking. Then the ``best'' anchor is selected to optimise the waypoint decoding during training as \textit{teacher forcing} \cite{teacher-forcing}. 

In this section, we discuss two types of frameworks named the \textit{predicted endpoint conditioned (PEC)} and \textit{prototype trajectory conditioned (PTC) frameworks} that use endpoints and prototype trajectories as anchors respectively.

\begin{table*}
\centering
\caption{Summary of predicted endpoint conditioned framework.}
\resizebox{\textwidth}{!}{
\begin{tabular}{ccccc} \toprule
PEC Model	& Publication & Endpoint Prediction      & Waypoints Prediction        & Mechanism for Scene Interaction \\ \midrule
PECNet \cite{pecnet} & ICCV 2020    & Coordinate Regression & Coordinate Regression & N/A \\
TNT  \cite{tnt} & CORL 2021   &  Coarse Heatmap Sampling + Offset      & Coordinate Regression  	& Heatmap Probability Prediction   \\ 
YNet  \cite{ynet} & ICCV 2021  &  Heatmap Sampling		& Heatmap Sampling  			& Heatmap Probability Prediction    \\
HOME  \cite{home}  & ITSC 2021 &  Heatmap Sampling		& Coordinate Regression 	 	& Heatmap Probability Prediction     \\  
TPNet \cite{tpnet} & CVPR 2020 & Coordinate Regression 	& Coordinate Regression 		& Proposal Refinement \\ 
DenseTNT \cite{densetnt} & ICCV 2021    & Coordinate Regression & Coordinate Regression & Pseudo Endpoint Supervision \\ 
ExpertTraj \cite{experttraj} & ICCV 2021 & Retrieval from Expert Repository & Coordinate Regression & Expert Selection \\
Gmtp \cite{goal-driven-prediction} & WACV 2021 & Region selection at Image Boundary & Coordinate Regression & Region Filtering based on Segmentation \\
GANet \cite{ganet} &  ICRA 2023   & Coordinate Regression & Coordinate Regression & GoICrop \\
\HL{CGTP} \cite{cgtp} &  \HL{TNNLS 2024}   & \HL{Pairwise Coordinate Regression} & \HL{Coordinate Regression} & \HL{Heatmap Probability Prediction} \\ 
\bottomrule  
\end{tabular}%
}
\label{tab:pec-framework}
\end{table*}

\noindent \textbf{Predicted Endpoint Conditioned (PEC) Frameworks.} Intuitively, agents can first decide the location where they will arrive and then plan their future trajectories \cite{goal-directed-prediction}. This leads to the \textit{PEC} framework where the endpoints can be predicted as an anchor and waypoints are generated to reach those positions. As shown in \cref{fig:pec}, this framework firstly predicts the endpoint distribution via an \textit{endpoint decoder}. Then, the \textit{waypoint decoder} predicts the middle locations given each selected endpoint. During training, the ground truth endpoint is selected so that the relationship between predicted waypoints and the conditioned endpoint is enhanced. During testing, the waypoint decoder uses $K$ endpoints sampled from the predicted endpoint distribution and generates waypoints for each of them. The PEC framework is widely used in current trajectory prediction methods due to its simplicity and effectiveness.
	
A summary of models using PEC frameworks can be seen in \Cref{tab:pec-framework}. PECNet \cite{pecnet} uses a CVAE to generate multiple endpoint coordinates. However, CVAE uses unexplainable latent variables to control the endpoints leading to the difficulty in integrating expert knowledge \cite{tnt}. To solve this problem, YNet \cite{ynet}, Goal-GAN \cite{goal-gan} and HOME \cite{home} directly predict an endpoint heatmap by integrating the observed trajectory and the scene segmentation image. These methods successfully integrate the scene information and provide scene-compliant endpoints. However, their performance highly relies on the resolution of the input images. TNT \cite{tnt} follows a similar way by (1) selecting multiple coordinates on a map as endpoint anchors; (2) predicting the probabilities on them; and finally, (3) estimating their offsets to the target endpoints. This method can effectively boost performance on low-resolution scene images. Moreover, it can be integrated with multiple kinds of data such as the lane endpoints in vehicle datasets. However, such models cannot make multiple predictions around the same endpoint candidates, which causes low performance when the candidate endpoints are not well selected \cite{densetnt}. To address this problem, DenseTNT \cite{densetnt} further performs cross-attention on features extracted from agents' trajectories and candidate endpoints, and regresses the coordinates of final endpoints. TPNet \cite{tpnet} proposes a two-stage PEC framework by first regressing the endpoint coordinates and then using a trainable module to filter and refine those waypoints to satisfy the terrain constraint. These methods do not suffer from the resolution problem and thus perform better. Meanwhile, ExpertTraj \cite{experttraj} suggests that endpoints can be obtained with a training-free process by sampling from existing trajectory repositories with minimal dynamic time-warping differences. Some recent studies propose to reduce the memory consumption of heatmap-oriented PEC framework \cite{ynet, goalsar, home} when dealing with multiple agents \cite{decouple-traj} or large amounts of sampling \cite{hypertraj}. Finally, a recent work, Goal Area Network (GANet) \cite{ganet}, models a goal area rather than coordinates to relax the strict goal prediction requirement, which ranked first on the leaderboard of Argoverse Challenge \cite{argoverse} in 2022. \HL{Finally,  CGTP \cite{cgtp} proposes the novel endpoint (goal) interactive prediction. Inspired by \cite{m2i}, this model integrates social interaction into endpoint predictions by conditionally assigning endpoint pairs to marginal and conditional agents.}

Moreover, the PEC framework can help the long-term prediction by conditioning on the endpoint and middle waypoints \cite{ynet,sgnet}. \cite{goal-driven-prediction} estimates the destination of the entire journey where the agent leaves the observed region to better control the future trajectory in a dynamic prediction horizon. Future PEC models can focus on avoiding unreachable endpoints due to the middle barrier and leveraging the diversity of waypoints to the same endpoint.

\begin{table*}[t]
\centering
\caption{Summary of predicted prototype conditioned framework.}
\begin{tabular}{c|ccc} \toprule
\textbf{Type}							              & \textbf{PTC Model}                    & \textbf{Prototype Trajectory Set}                                         \\ \midrule

\multirow{2}{*}{Clustering-based} 			 & MultiPath \cite{multipath} (CoRL2019), SICNet \cite{sicnet} (ICCV2023)   & Training set clustering (Large Datasets)                           \\
                                         & CoverNet  \cite{covernet} (CVPR2020)                & Training set bagging \\ 
                                         
                                         & \cite{eigentrajectory} (CVPR2023), \cite{singular-trajectory} (CVPR2024) & Singular space clustering \\
                                         \midrule

\multirow{3}{*}{Physics-based}          & S-Anchor  \cite{social-anchor} (CVPR2021)               & Discrete Choice Model \\
                                         & MultiPath \cite{multipath} (CoRL2019) & Different constant velocity motions (Small Datasets) \\
                                         & SIT      \cite{social-interpretable-tree} (AAAI2022)                  & Tree-like Paths                                    \\ \midrule
\multirow{2}{*}{Learning-based}            & PCCSNet   \cite{pccsnet} (ICCV2021)         & 	Conditional generation                                          \\ 
                                         & \begin{tabular}{m{8cm}}\centering \cite{mmtransformer} (CVPR2021), \cite{multipath-pp} (ICRA2022), \cite{qcnet,prophnet} (CVPR2023)  \end{tabular}
                                         & 	DETR \cite{detr}-like Query-based Generation		
                                        								 \\
                                         \bottomrule

\end{tabular}%
\label{tab:ptc-framework}
\end{table*}

\noindent \textbf{Prototype Trajectory Conditioned (PTC) Frameworks.}
Prototype trajectories are typical types of motion patterns indicating how the agents might follow in the future. For example, agents turning left all share a curved motion pattern towards to left and a trajectory satisfying this pattern can be a prototype trajectory. The PTC framework learns to select suitable candidates from the prototype trajectory set and refine them based on agents' surrounding environments and their motion states.  As shown in \cref{fig:ptc}, prototype trajectories and agents' current motion states are encoded through the prototype and past encoders. Then, they are sent into the prototype selector to rank the suitable prototype trajectories. Finally, the \textit{refiner} generates the final outputs based on features from the observed and prototype trajectories.

To build the anchor set with sufficient diversity, prototypes can be collected in different ways. A summary of models following PTC framework can be seen in  \Cref{tab:ptc-framework}. Some studies use \textit{clustering}-based methods \cite{multipath,covernet} to build prototypes using the existing datasets. For example, MultiPath \cite{multipath} and SICNet \cite{sicnet} \textit{cluster} trajectories in existing datasets using the \textit{k}-means algorithm while \cite{eigentrajectory, singular-trajectory} cluster trajectories in a singular space. CoverNet \cite{covernet} uses a ``bagging'' algorithm to greedily select candidate trajectories that can cover the whole training set. Moreover, some \textit{physics}-based methods manually construct prototypes, which are generated using physical models. For example, MultiPath \cite{multipath} suggests to generates straight trajectories with different directions and speeds as anchors when small datasets are given. S-Anchor \cite{social-anchor} constructs the set with different levels of speeds and directions via a discrete choice model \cite{discrete-choice-model} to integrate social interactions. SIT \cite{social-interpretable-tree} builds a tree-like route map and dynamically selects and refines the path segments.
Finally, to encourage more complex prototypes, some methods choose \textit{learning}-based algorithms to generate prototypes using deep learning models. PCCSNet \cite{pccsnet} clusters all trajectory features into several groups and uses cluster centres to generate prototype trajectories. Recently, some methods such as mmTransformer \cite{mmtransformer}, MultiPath++ \cite{multipath-pp}, QCNet \cite{qcnet} and ProphNet \cite{prophnet} follow DETR \cite{detr} to generate diverse prototypes using queries with a modality-agnostic manner and each query embedding corresponds to a prototype trajectory.  As suggested in \cite{qcnet}, this allows the queries to collaborate with each other via the self-attention mechanism when generating trajectory proposals. 

Finally, there are multiple designs for \textit{refiners}. Most methods \cite{multipath, qcnet, multipath-pp, social-anchor,prophnet} predict residuals and aggregate them on prototypes for final predictions while PCCSNet \cite{pccsnet} directly generates final predictions and SICNet \cite{sicnet} masks prototype trajectories and complete them.

\subsection{Recurrent-based MTP Framework}
\label{sec:rnn-mtp-framework}
This framework refers to the models that use auto-regressive decoders to generate future trajectories using recurrent neural networks, where the predicted location at the current time step is conditioned on the previous sampled locations and the internal states at previous time steps. In other words, sampling different locations at each time step encourages the diversity of the distribution of all future trajectories. Therefore, models following this framework should recurrently predict the distribution of all possible locations at each time step and sample multiple locations for future steps. Finally, $K$ entire trajectories are obtained from all predictions. Here, we review three categories of sampling strategies:

\noindent \textbf{Gaussian Sampling.}  One solution is to use a bivariate Gaussian to describe the position distribution \cite{social-lstm,trajectron,trajnet++}. Instead of regressing the exact 2D coordinates, models predict the Gaussian parameters at each time step including the mean and standard deviation of the 2D locations and a correlation value. Sampled 2D locations further become one of the inputs for the next prediction. However, the main drawback of this strategy is that bivariate Gaussian is too simple to describe the complex distribution of future locations.

\noindent \textbf{Grid Sampling.} An alternative way is to employ occupancy grid maps to indicate which location the agent will go to in the next time step. As shown in \cref{fig:grid}, the scene is divided into grid cells and the model predicts the occupancy probability in each cell determined by observed information for each time step. Then multiple possible positions at each time step are categorically sampled (non-differentiable) \cite{multiverse}. In addition, Gumbel-Softmax can be used for differentiable sampling \cite{tdor} during the model training. The main benefit of the grid-based outputs is that they can be highly compliant with scenes with advanced training strategies such as reinforcement learning or occupancy losses and are suitable for long-term prediction. However, it is rarely used due to significant computation from the convolutional operations and high sensitivity to the resolution of the maps.

\noindent \textbf{Token Sampling.} A recent study LMTraj \cite{lmtrajectory} proposes a language model for trajectory prediction, where predicted trajectories are in the form of word tokens. Therefore, at each future timestep, LMTraj can sample different positions by modulating the word token probability. LMTraj has demonstrated outperformance on multiple datasets for both STP and MTP, indicating that it can be a new trend in MTP with the development large language model.

With all types of outputs, during the inference, there is a need to obtain $K$ trajectories from all predictions for diverse prediction. The simplest way is to randomly select $K$ different trajectories. However, this cannot encourage the diversity of predicted distribution. A more advanced way in Multiverse \cite{multiverse} and ST-MR \cite{stmr} is to use beam search \cite{beam-search} for trajectories with top $K$ accumulated log probabilities. Alternatively,  we can use a sampling trick in \cref{sec:sampling-trick} to diversely sample $K$ trajectories from an oversampled prediction set.

\subsection{Other Techniques for Improved MTP}
\noindent \textbf{Variety Loss} To generate diverse predictions, we can optimise the model using the \textit{variety loss} \cite{social-gan}, also named \textit{Winner-Takes-All Loss} in \cite{tpa,ewta}, using the minimum reconstruction error: 
\begin{equation}
    L_{variety}(\hat{Y}_{K,i}^\tau,Y_i^\tau) = \min_{k<K} L_{rec}(\hat{Y}_{k,i}^\tau, Y_{i}^\tau).
\end{equation}
As mentioned in \cite{social-gan}, variety loss can effectively alleviate the mode collapse problem brought from the reconstruction loss. This strategy is commonly used in noise-based MTP frameworks \cite{stgat,tpa, ewta, tp-raster}.  A study \cite{variety-loss-analysis} further explains that a learner trained with variety loss can converge to the square root of the ground truth probability density function.

\noindent \textbf{Bivariate Gaussian for Outputs Representation.}
\label{sec:bivaiate-gaussian}
The bivariate Gaussian for outputs representation for MTP introduced in \cref{sec:rnn-mtp-framework} can also be an enhancement of MTP prediction. This strategy was first used in S-LSTM \cite{social-lstm} for deterministic prediction but was depreciated by GAN-based models due to its non-differentiable position sampling. Then, it is reused in Social-STGCNN \cite{stgcnn} and \cite{sgcn} for MTP where multiple trajectories can be obtained by sampling $K$ future positions from the predicted distributions. However, the output positions are sampled individually and may not be temporally correlated, causing unrealistic predictions. 

We find that in most studies, the bivariate Gaussian can be used to indicate the entropy of predicted distribution or relaxation of predictions for the loss calculation. For example, it can be combined with anchor-based MTP frameworks to avoid the expectation-maximisation training procedure and visualise the uncertainties under each time step for better optimisation \cite{multipath}. 

\noindent \textbf{Diverse Sampling Tricks.}
\label{sec:sampling-trick}
Random sampling from an MTP model may not cover all modes due to limited sampling numbers. Therefore, \textit{diverse sampling tricks} are proposed to ensure the coverage of the distribution. For example, Likelihood Diverse Sampling (LDS) 
 \cite{likelihood-diverse-sampling} is proposed as a \textit{post-hoc} method to enhance the quality and diversity by training a sampling model to balance the likelihood of an individual trajectory and the spatial separation among trajectories. \cite{ynet} proposes the Test Time Sampling Trick to cluster sampled endpoints into $K$ centres for wider coverage of predicted endpoints. TDOR \cite{tdor} sends an oversampled trajectory set into a transformer to generate $K$ trajectories, where its generative style decoder can be considered as a learnable cluster algorithm using self-attention. Non-Probability Sampling Network (NPSN) \cite{npsn} is a Quasi-Monte Carlo method to generate non-biased samples. Stimulus Verification \cite{stimulus-verification} filters predicted samples by verifying their compliance with scene and social interaction. BOSampler \cite{bosampler} tackles the long-tail problems in Monte-Carlo-based sampling by sequentially exploring the predictions in the long-tail region based on a Gaussian Process. 

\noindent \textbf{Joint Sampling Tricks.} Most MTP models sample the future trajectories for each agent independently at test time without considering their future interactions. To tackle this problem, \textit{joint sampling tricks} are proposed to sample trajectories for all agents collaboratively. For example, noise-based models such as S-GAN \cite{social-gan} predict future trajectories for all agents with shared noises. This method predicts trajectories for all agents in a non-autoregressive manner. Another method is to predict the trajectories with priorities, where agents with higher priority predict the trajectories first followed by agents with lower priority. For example, \cite{tp-fictitious-play} uses the fictitious play \cite{fictitious-play} to model the interplay among crowds. \cite{multiple-futures-prediction} rollouts the joint predictions via autoregressive decoding, where the agents' trajectories are predicted one by one. ScePT \cite{scept} fixes the trajectories of conditioned agents and predicts trajectories for the remaining ones. Similarly, M2I \cite{m2i} is an effective joint sampling trick for pairwise interactions. It splits all agents into influencers and reactors pairwisely, where reactors decide their routes based on influencers' behaviours. CGTP \cite{cgtp} extends M2I to focus on handling the interactions on endpoints. IPCC-TP \cite{ipcc-tp} models pairwise joint Gaussian distributions to simulate the interactions among agents.

\begin{table*}[t]
    \centering
    \caption{ Details about real-world datasets for MTP.}
    \begin{tabular}{c|ccccc} 
    \toprule
	 Name & View & Annotations & \#Agents & \#Locations & Agent Types  \\\midrule
	 \multicolumn{6}{c}{\scriptsize{Commonly Used Datasets}} \\ \midrule
	 ETH\&UCY \cite{eth,ucy} & CCTV (Urban/Campus) & Trajectories, Videos & 2205 & 5 & Pedestrians \\
	 Stanford Drone \cite{sdd} & Bird Eye (Campus) & Trajectories, Videos & 10.3K & 8 & Pedestrians, Vehicles \\
	 InD \cite{ind} & Bird Eye (Urban)  & Trajectories, Images, HD Maps & 13.6K & 4 & Pedestrians, Vehicles \\
	 NuScenes \cite{nuscenes}  & Bird Eye (Urban) & Trajectories, HD Maps &  - & 2 & Pedestrians, Vehicles \\
	 Argoverse \cite{argoverse}  & Bird Eye (Urban) & Trajectories, HD Maps &  11.7M & 2 & Pedestrians, Vehicles \\
	 Waymo \cite{waymo}  & Bird Eye (Urban) & Trajectories, HD Maps & 7.6M & 6 & Pedestrians, Vehicles \\
	
	 Interaction \cite{interaction} & Bird Eye (Urban)  & Trajectories, HD Maps & 40K & 11 & Vehicles \\ \midrule
	 \multicolumn{6}{c}{\scriptsize{Datasets with Annotated Intentions}} \\ \midrule
	 TrajNet++ \cite{trajnet++} & Mixed & Trajectories, Intentions & - & - & Pedestrians \\
	 VIRAT \cite{virat} & CCTV (Street/Car Park)  & Trajectories, Videos, Intentions & 91.3K & 8 & Pedestrians, Vehicles \\
	 PANDA \cite{panda} & CCTV (Streets/Campus)   & Trajectories, Videos, Intentions & 12.7k & 15 & Pedestrians \\  
	 PIE \cite{pie} & First Person (Urban)   & Trajectories, Videos, Intentions & 1.8K & - & Pedestrians \\ 
	 JAAD \cite{jaad} & First Person (Urban)  & Trajectories, Videos, Intentions  & 2.8K & - & Pedestrians \\
		
	 TITAN \cite{titan} & First Person (Urban)  & Trajectories, Videos, Intentions  &  39.6K & - & Pedestrians, Vehicles \\
	 Loki \cite{loki} & First Person (Urban)  & Trajectories, Videos, Intentions  &  28K & - & Pedestrians, Vehicles \\ \midrule
	 
	\multicolumn{6}{c}{\scriptsize{Dataset with Annotated Multiple Future Trajectories}} \\ \midrule
	ForkingPath \cite{multiverse} & \begin{tabular}{m{3.0cm}}\centering  Simulated Multiview\\(Street,Car Park)  \end{tabular}  &  \begin{tabular}{m{3cm}}\centering Trajectories, Videos, Multi-path Annotations \end{tabular}  & 2.6K & 7 & Pedestrians \\ \bottomrule

    \end{tabular}
    \label{tab:datasets}
\end{table*}

\section{Datasets for MTP}
\label{sec:datasets}
This section discusses three types of datasets used in MTP. The details of real-world datasets are summarised in \Cref{tab:datasets}.

\subsection{Commonly Used Datasets} 
\label{se:common-data}
Commonly used datasets are those captured from the real world and used in both STP and MTP by default. Pedestrian trajectory prediction studies mostly use ETH\&UCY \cite{eth,ucy} and Stanford Drone Dataset (SDD) \cite{sdd} while vehicle trajectory prediction often uses InD \cite{ind}, NuScenes \cite{nuscenes}, Argoverse \cite{argoverse}, INTERACTION \cite{interaction} and Waymo \cite{waymo} datasets. Observed and ground truth (future) trajectory samples are obtained via sliding windows on entire sequences. The environment information is presented by videos, reference images or high-definition (HD) maps. However, these datasets provide only one ground truth for each agent. Therefore, the training and evaluation procedures are usually similar to STP. 

\subsection{Datasets with Annotated Intentions} 
There are also trajectory prediction datasets containing intention labels annotated by human experts. Trajnet++ \cite{trajnet++} is a large-scale interaction-centric trajectory-based benchmark that concatenates multiple real-world pedestrian trajectory prediction datasets. Specifically, it defines multiple pedestrian behaviour categories such as leader following, collision avoidance and grouping and provides algorithms to classify them. VIRAT \cite{virat}, PANDA \cite{panda}, PIE \cite{pie}, JAAD \cite{jaad} and TITAN \cite{titan} contain the trajectories with rich activities including moving (fast/slow), stopping, crossing, etc.  LOKI \cite{loki} provides long-term and short-term intentions. 

As mentioned, agents' intentions can be considered as modalities for MTP. Therefore, datasets providing annotated intentions can be used to guide the model training and enhance the evaluation. For example, we can force the model to select the modalities among provided intentions and specifically compare the predicted trajectories under the annotated intentions with the ground truth future path.

\subsection{Datasets with Annotated Multiple Future Trajectories}
\label{sec:dataset-multifuture}
\noindent \textbf{Synthetic Toy Datasets.} The distribution of trajectories in each dataset is implicit and hence it is difficult to evaluate whether the model correctly fits the distribution. Therefore, synthetic datasets have been proposed with simple and controllable distributions for evaluation. For example, \cite{socialways} proposes a toy dataset with six groups of trajectories, each group starting from one specific point and following three different ways to the endpoint. \cite{multipath} proposes a 3-way intersection toy dataset with the probability of choosing the left, middle or right path set. Experiments using these datasets highlight the problems of mode collapse and social-acceptance in current frameworks.

\noindent \textbf{ForkingPath.}
Liang \textit{et al.} \cite{multiverse} suggest that current trajectory prediction datasets only provide one possible ground truth trajectory for each agent, which is not suitable for MTP evaluation. One solution is to manually annotate extra future trajectories. However, this requires huge effort and the trajectories may not be realistic due to the single view. To tackle this problem, Liang \textit{et al.} \cite{multiverse} simulate the real-world data using CARLA \cite{carla} and let multiple human experts control the pedestrians from multiple views. Finally, multiple ground truths can be collected to construct the dataset named ForkingPath. Further studies \cite{mggan-tp,likelihood-diverse-sampling} have used this dataset to compare the predicted and ground truth distributions. 
 \begin{figure*}[t]
    \centering
     \begin{subfigure}[b]{0.24\textwidth}
         \centering
         \includegraphics[width=\textwidth]{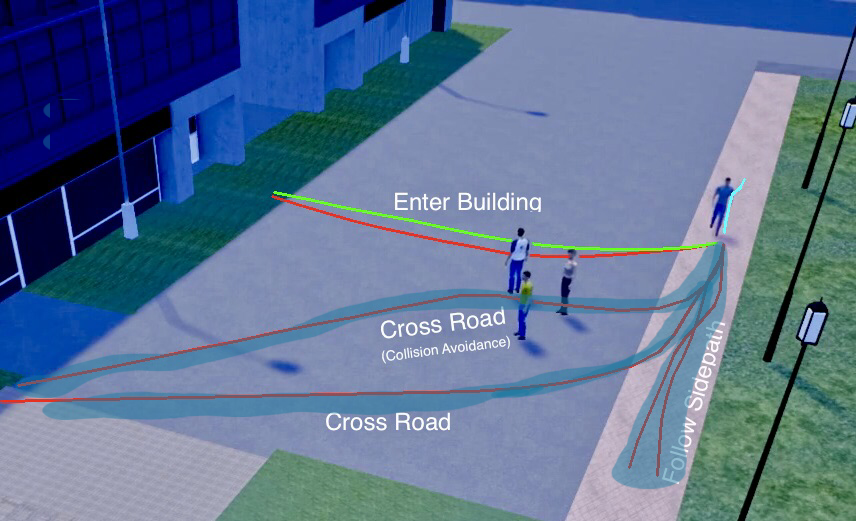}
         \caption{Acc($\times$) Div($\checkmark$) Soc($\checkmark$)}
         \label{fig:diverse-acc}
     \end{subfigure}
    \begin{subfigure}[b]{0.24\textwidth}
         \centering
         \includegraphics[width=\textwidth]{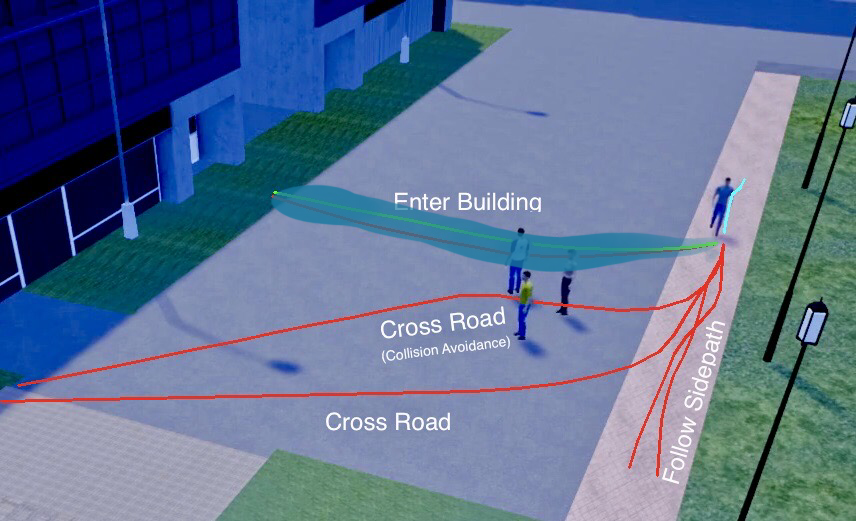}
         \caption{Acc($\checkmark$) Div($\times$) Soc($\checkmark$)}
         \label{fig:acc-diverse}
     \end{subfigure}
    \begin{subfigure}[b]{0.24\textwidth}
         \centering
         \includegraphics[width=\textwidth]{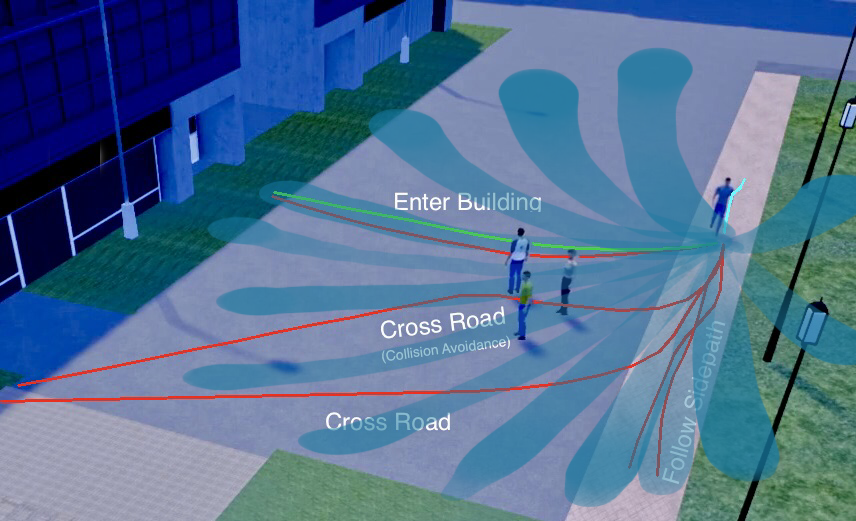}
         \caption{Acc($\checkmark$) Div($\checkmark$) Soc($\times$)}
         \label{fig:diverse-acceptable}
     \end{subfigure}
     \begin{subfigure}[b]{0.24\textwidth}
         \centering
         \includegraphics[width=\textwidth]{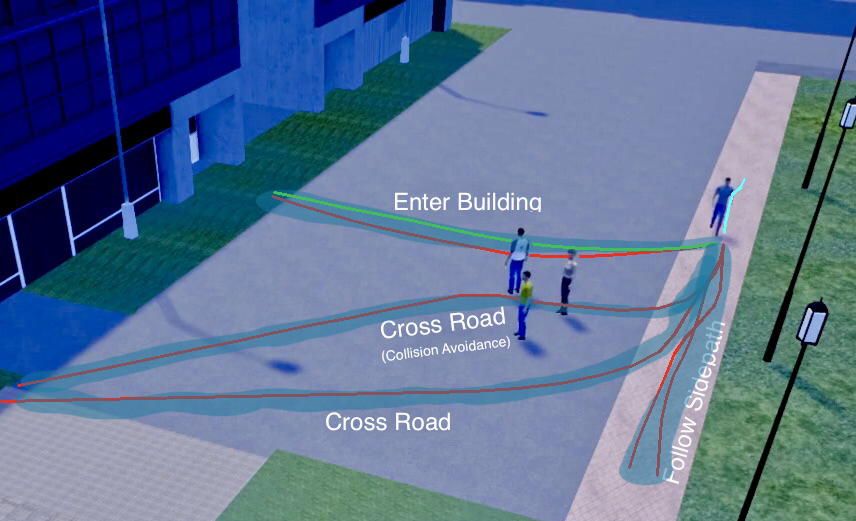}
         \caption{Acc($\checkmark$) Div($\checkmark$) Soc($\checkmark$)}
         \label{fig:expected}
     \end{subfigure}

     \caption{Examples of MTP satisfying accuracy (Acc), diversity (Div) and social-acceptance (Soc), where the shaded part denotes the predicted distributions. Case studies: (a) Predictions do not cover the ground truth. (b) Predictions hit the ground truth but only cover a single mode. (c) Predictions cover the ground truth and are diverse, but some of them enter the non-walkable zone. (d) The expected predictions.}
     \label{fig:mtp_expect}
\end{figure*}
\begin{figure*}[t]
    \centering
    \resizebox{0.95\textwidth}{!}{
	\begin{forest}
  for tree={
      grow=east,
      reversed=true,
      anchor=base west,
      parent anchor=east,
      child anchor=west,
      base=left,
      font=\small,
      rectangle,
      draw,
      rounded corners,align=left,
      minimum width=2.5em,
      inner xsep=4pt,
      inner ysep=0.5pt,
  },
  where level=1{fill=blue!10,align=center}{},
  where level=2{font=\footnotesize,fill=pink!30,align=center}{},
  where level=3{font=\footnotesize,yshift=0.26pt,fill=yellow!20}{},
        [Evaluation Metrics , fill=gray!20
          [Lower-Bound based Metrics
              [\emph{MoN} [minADE$_K$; minFDE$_K$]
              ]
              [\emph{JMoN} [minJADE$_K$;minJFDE$_K$ \cite{joint-metric-matters, waymo}]
              ]
              [\emph{MR} [MR in Argoverse \cite{argoverse}; Waymo \cite{waymo}; \\
              NuScenes \cite{nuscenes}]
              ]
          ]
          [Probability-aware Metrics
            [\emph{ML}
                [ML + ADE/FDE \cite{pcmd}; ML+Overlap Rate \cite{waymo}; \\ mAP/Soft mAP \cite{waymo} ]
            ]
            [\emph{TopK}
                [PCMD \cite{pcmd}; MoN + Sampling Tricks]
            ]
            [\emph{Gaussian-based Metrics}
                [KDE-NLL \cite{trajectron}; AMD/AMV \cite{social-implicit}]
            ]
          ]
          [Distribution-aware Metrics 
            [\emph{Coverage}[EMD \cite{socialways}; Precision \& Recall \cite{mggan-tp}; PTU \cite{stmr},fill=yellow!20]]
          ]
        ]
\end{forest}
    }
    \caption{An overview of the taxonomy of MTP evaluation metrics.}
    \label{fig:good-mtp}
\end{figure*}
\section{Evaluation Metrics}
\label{sec:evaluation-metrics}
\subsection{Preliminary: Evaluation Metrics for STP}
In STP, the default evaluation metrics are \textit{average displacement error (ADE)} and \textit{final displacement error (FDE)}, which measure the $l_2$ distances to ground truth trajectories through all future timesteps and the last future timestep respectively:
 \begin{align}
     ADE &= \frac{1}{N * T_{pred}} \sum_{i<N, t\le T_{pred}} ||\hat{Y}_i^t -  Y_i^t||_2 \\
     FDE &= \frac{1}{N} \sum_{i<N,t=T_{pred}} ||\hat{Y}_i^t -  Y_i^t||_2
 \end{align}
In addition, there are multiple metrics examining social acceptance such as \textit{collision rate}, \textit{overlap rate} and \textit{offroad rate} which measure whether the predicted trajectory of the agent collides with surrounding agents or enters inaccessible regions.

Evaluation metrics for MTP need to take all predictions into consideration and thus are more challenging where models are expected to provide \textit{diverse}, \textit{accurate} and \textit{socially-acceptable} predictions as shown in \cref{fig:mtp_expect}. In this section, we review these MTP metrics with the taxonomy in \cref{fig:good-mtp} and summarise their advantages and disadvantages in \Cref{tab:summary-eval}.

\subsection{Lower-bound-based Metrics} 
\label{sec:lower-bound}
Lower-bound-based metrics aim to evaluate whether the ground truth can be covered by the prediction. Given $K$ predicted trajectories, each prediction is compared with the ground truth and the best score is recorded without considering the exact confidence. Therefore, these metrics can be simply converted from those used for STP and are valid for any models in trajectory prediction. 

\noindent \textbf{Minimum-of-N (MoN).} MoN is first proposed in \cite{social-gan} and is the default metric for most MTP works. It calculates the minimum error among all predictions:

\begin{equation}
    MoN = \mathbb{E}_{i,t\in \tau} \min_{k<K} DE(\hat{Y}_{i,k}^t, Y_{i}^t) 
\end{equation}
where $DE$ can be any distance metric used in STP. 
Many studies adopt this strategy to adapt ADE and FDE for multimodal prediction, abbreviated as \textit{minADE$_K$} and \textit{minFDE$_K$} and they have become the default metrics in all multimodal trajectory prediction methods and benchmarks. 

\noindent \textbf{Joint Minimum-of-N (JMoN).} To consider social interaction during the evaluation, some studies introduce JMoN (e.g., minJ(A/F)DE$_K$) \cite{social-gan,sgan-v2,joint-metric-matters,m2i,waymo} which calculates the average performance for all observed agents and then select the best of them:
\begin{equation}
    JMoN =  \min_{k<K} \mathbb{E}_{i,t\in \tau} DE(\hat{Y}_{i,k}^t, Y_{i}^t) 
\end{equation}
In other words, JMoN conducts the MoN at a scene-level and thus is able to measure the social interaction among agents. However, as mentioned in \cite{stgcnn}, JMoN is not applicable to some MTP frameworks. For example, some PEC models \cite{ynet,pecnet} sample endpoints without considering the neighbour agents while some PTC models select their prototypes independently. Therefore, JMoN is not widely used in MTP.

\noindent \textbf{Miss Rate (MR).} Some vehicle trajectory prediction benchmarks such as Waymo, Argoverse and nuScenes use MR to indicate whether the ground truth can be covered by the predictions. A prediction \textit{misses} the ground truth if it is more than $d$ metres from the ground truth according to their displacement error and \textit{hits} otherwise. MR counts the scenarios that all predictions miss the ground truth:

\begin{equation}
    MR = \mathbb{E}_{i,t \in \tau} \sign((\min_{k<K} DE(\hat{Y}_{i,k}^t, Y_{i}^t))) - d
\end{equation}
where FDE is used as the displacement metric in Argoverse and Waymo benchmarks and ADE is used in NuScenes. The distance threshold $d$ is 2 meters in Argoverse and NuScenes and is adjusted with speed in Waymo benchmark.

\noindent \textbf{Challenge: Information Leak.} 
Lower-bound-based metrics are sensitive to randomisation and are insufficient indicators of the performance of models. Information leaks happen during testing since only the best prediction is used for evaluation based on the distances to the ground truth. This allows a distribution with high entropy for a lower error. For example, the constant velocity model \cite{cvm} may even ``outperform'' deep learning-based models by adjusting the angles for a wider spread distribution. This further results in an unreliable indication of interaction handling. For example, a prediction that violates the social rules can be generated from an STP model without social interaction modules and neglected in MTP since the best one is selected.

\begin{table*}[t]
\centering
\caption{Summary of MTP evaluation metrics.}
\begin{tabular}{l|ccc|c|c}\toprule
         Metrics     & Acc & Div & Soc        & Advantages                                                                                              & Disadvantages                                                  \\\midrule \midrule

Lower-bound-based Metrics &$\checkmark$  && & \multicolumn{1}{m{0.3\textwidth}|}{$\bullet$ Measures the diversity of the predicted trajectories. $\bullet$ Can be easily implemented based on any STP metrics;} & \multicolumn{1}{m{0.3\textwidth}}{$\bullet$ Sensitive to randomisation. $\bullet$ Sensitive to the number of allowed predictions; $\bullet$ Allows unacceptable predictions.}                                  \\\midrule
Probability-aware Metrics & $\checkmark$ & & & \multicolumn{1}{m{0.3\textwidth}|}{$\bullet$ Considers the probability of each prediction. $\bullet$ Less sensitive to randomisation.} & \multicolumn{1}{m{0.3\textwidth}}{$\bullet$ Fail to measure the diversity of predictions. $\bullet$ Requires huge number of predictions to indicate the distribution if the probabilities are not provided.}   \\\midrule
Distribution-aware Metrics  &$\checkmark$ &$\checkmark$&$\checkmark$ & \multicolumn{1}{m{0.3\textwidth}|}{$\bullet$ Captures the manifold of distributions between the ground truth and predictions.}                     & \multicolumn{1}{m{0.3\textwidth}}{$\bullet$ Need human annotations for multiple ground truth trajectories.} \\\midrule
\end{tabular}%
\label{tab:summary-eval}
\end{table*}
\subsection{Probability-aware Metrics}
Probability-aware metrics measure how likely the ground truth can be sampled from the predicted distribution. In contrast to lower-bound metrics, MTP models are required to assign the highest probability to the best prediction.

\noindent \textbf{Most-likely (ML) based Metrics.} The simplest way is to select the prediction with the highest probability to perform the STP evaluation. For example, the ML metric \cite{pcmd} simply selects the most likely prediction for ADE and FDE calculation, as well as the overlap rate calculation in the Waymo benchmark. Similarly, the \textit{mean average precision (mAP)} is used in the Waymo benchmark \cite{waymo}. The most likely prediction is considered a true positive if it aligns with the ground truth; otherwise, it is a false positive. All other predictions are assigned a false positive. Then, it computes the area under the precision-recall curve. From 2022, Waymo benchmark uses \textit{Soft mAP}, which is the same as mAP except that it ignores the penalty from the predictions other than the most likely one. Recently, \cite{mon-auc} proposes a new evaluation metric named Area Under the Curve, which randomly samples $K$ predictions to calculate the expectations of the minADE.

\noindent \textbf{TopK based Metrics.} \cite{pcmd} suggests that one prediction cannot represent the whole distribution. Therefore, we can select candidates with a probability larger than a threshold $\gamma$ among $M\gg K$ predictions for MoN evaluation, known as the \textit{probability
cumulative minimum distance (PCMD)}:
\begin{equation} 
    \text{PCMD} = MoN(\hat{Y}_{i,k} | \mathcal{P} (\hat{Y}_{i,k'} | X_i^T, k'< M) \ge \gamma)
\end{equation}
Then, predictions with top $K$ probabilities are selected. 
However, it cannot be used if the probability of each prediction is not provided. To tackle that, we can select $K$ predictions using sampling tricks in \cref{sec:sampling-trick}.

\noindent \textbf{Gaussian based Metrics.} If no probability is provided, an alternative method is to first estimate a Gaussian distribution given $K$ discrete predictions using a method such as \textit{kernel density estimation} (KDE), by estimating the probability density function given a sequence of independent random variables. In trajectory prediction, \cite{trajectron} firstly introduces KDE-NLL as one of the evaluation metrics in MTP, which computes the mean log-likelihood of the ground truth trajectory for each future time step:
\begin{equation}
    \text{KDE-NLL} = - \mathbb{E}_{i,t\in \tau} \log \mathcal{P}(Y_{i}^t |  \text{KDE}(\hat{Y}_{i,K}^t)),
\end{equation}
and is further used in subsequent studies such as \cite{pecnet,ynet}. \cite{social-implicit} further improves KDE-NLL by proposing \textit{Average Mahalanobis Distance} (AMD), which measures the distance between the ground truth and the generated distribution, and \textit{Average Maximum Eigenvalue} (AMV) to measure the confidence of the predictions.

\noindent \textbf{Challenge: Ground Truth May Not Be Most Likely.}
Probability-aware metrics can suffer more from noisy datasets. Some of predictions can be more reasonable than the ground truth based on observed clues and thus should not be penalised. For example, if an agent has a behaviour such as ``zig zag'' in pedestrian datasets and ``sudden cut in'' in vehicle datasets, how likely should this behaviour happen? In these cases, we believe that lower-bound based metrics are more suitable.

\subsection{Distribution-aware Metrics}
\label{sec:metrics-distribution}
As explained in \Cref{tab:summary-eval}, none of the metrics above evaluates the topology of the distribution of all predicted trajectories. Specifically, they do not penalise socially unacceptable predictions or outside the ground truth distribution, mainly because only one ground truth is provided and the real ground truth distribution cannot be estimated. To address this issue, datasets (e.g., ForkingPath) that provide multiple ground truth trajectories for each agent are constructed to measure the coverage between the predicted and ground truth distributions.
 
\noindent \textbf{Coverage-based Metrics.} \cite{socialways} propose the \textit{earth-moving distance} (EMD) by calculating the ADE results with linear sum assignment between predicted and ground truth samples. \cite{mggan-tp} proposes the \textit{recall} and \textit{precision} metrics for generative networks to measure the coverage. Given the predicted and real future trajectory set, \textit{recall} counts how many predicted trajectories can find a ground truth trajectory that lies inside a certain range $d$.
\begin{equation}
    Recall = \mathbb{E}_{k<K_G} (\min_{k'< K_R} || \hat{Y}_{i,k}^t - Y_{i,k'}^t ||_2 ) < d
\end{equation}
where $K_G$ is the number of predictions and  $K_R$ is the number of annotated ground truths for agent $i$. In other words, the predicted distribution should cover all ground truth trajectories. On the other hand, the \textit{precision}, calculates the ratio of generated samples in support of the ground truth distribution and penalises out-of-distribution predictions: 
\begin{equation}
    Precision = \mathbb{E}_{k<K_R} (\min_{k'< K_G} || Y_{i,k}^t - \hat{Y}_{i,k'}^t ||_2 ) < d
\end{equation}
A metric similar to \textit{precision} is named \textit{Pecentage of Trajectory Usage} (PTU) \cite{stmr}, which measures the ratio of annotated future trajectories that have been assigned to the best prediction during the minADE$_k$ and minFDE$_k$ calculations.

\noindent \textbf{Challenge: Heavy Annotation Effort.} 
Distribution-aware metrics require extra annotations and corrections by human experts on real-world datasets which is labour-intensive. Also, even human annotation cannot guarantee the coverage of all modalities. Although synthetic datasets can alleviate this problem, they can only evaluate simple and unrealistic interactions. Therefore, these metrics are not used in most benchmarks.

\begin{table*}
    \caption{Performance (minADE$_{20}$/minFDE$_{20}$) of MTP models on pedestrian trajectory prediction benchmarks.}
    \label{tab:comparing-eth-ucy-sdd}
    \centering
    \begin{tabular}{l|c|c|c|c|c} \toprule
     Frameworks & Model & Publication & ETY\&UCY & SDD  & Source of Results \\ \midrule
     \midrule
     Noise (GAN) & SGAN \cite{social-gan} & CVPR 2018 & 0.53/1.04 & 13.58/24.59 & \cite{npsn} \\ 
     Noise (GAN) & MG-GAN \cite{mggan-tp} & ICCV 2021  & 0.37/0.71 & N/A & \cite{mggan-tp} \\ 
     Noise (VAE) & Trajectron++ \cite{trajectron-pp} & ECCV 2020 & 0.31/0.52  &  11.40/20.12 & \cite{npsn} \\
     Noise (VAE) & Agentformer \cite{agentformer} & ICCV 2021 &  0.26/0.44  & N/A & Reproduced \\
     Noise (VAE) & SocialVAE \cite{social-vae} &ECCV 2022 & 0.24/0.42  & 8.88/14.81 & \cite{social-vae} \\
     Noise (Diffusion) & MID \cite{mid} &CVPR 2022 & 0.31/0.53 & 7.61/14.30 & \cite{mid}  \\
     Noise (NF) & FlowMo \cite{flomo} & IROS 2021 & 0.22/0.37 & N/A & \cite{flomo} \\ 
     Noise (NF) & STGlow \cite{stglow} & TNNLS 2023  & \textbf{0.15/0.28} & \textbf{7.20}/11.20 & \cite{stglow} \\ \midrule
     Anchor (PEC) & PECNet \cite{pecnet} & ECCV 2020  & 0.32/0.56  & 9.97/15.89 & \cite{pecnet} \\ 
     Anchor (PEC) & YNet \cite{ynet} & ICCV 2021 & N/A & 8.97/14.61 & \cite{mid} \\ 
     Anchor (PTC) & PCCSNet \cite{pccsnet} & ICCV 2021 & 0.21/0.43 & 8.62/16.61 & \cite{pccsnet} \\
     Anchor (PTC) & SIT \cite{social-interpretable-tree} & AAAI 2022 & 0.23/0.38 & 9.13/15.42 & \cite{social-interpretable-tree} \\\midrule
     Recurrent (Gaussian) & Trajectron++ \cite{trajectron-pp} & ECCV 2020 & 0.31/0.52  &  11.40/20.12 & \cite{npsn} \\
     Recurrent (Grid) & TDOR \cite{tdor} & CVPR 2022 & N/A & 8.78/14.34 & \cite{tdor}\\ 
     Recurrent (Token) & LMTraj \cite{lmtrajectory} & CVPR 2024 & 0.22/0.32 & 7.8/\textbf{10.1} & \cite{lmtrajectory}\\ \midrule
     Diverse Sampling & YNet + TTST \cite{ynet} & ICCV 2021  & 0.18/\textbf{0.27} & 7.85/11.85 & \cite{ynet} \\
     Diverse Sampling & TDOR + TTST \cite{tdor} & CVPR 2022  & N/A & 7.64/12.12 & \cite{tdor} \\
     Diverse Sampling & TDOR + GSD \cite{tdor} & CVPR 2022  & N/A & \textbf{6.77}/10.46 & \cite{tdor} \\
     Diverse Sampling & SocialVAE + TTST \cite{social-vae} & ECCV 2022 & 0.21/0.33 & 7.64/12.12 & \cite{social-vae} \\
     Diverse Sampling & AgentFormer + DLow \cite{agentformer} & ICCV 2021 & 0.23/0.39 & N/A & \cite{agentformer} \\
     Diverse Sampling & AgentFormer + LDS  \cite{likelihood-diverse-sampling} &ICCV 2021 & 0.24/0.36 & N/A & \cite{stimulus-verification} \\
     Diverse Sampling & AgentFormer + SV \cite{stimulus-verification} & CVPR 2023 & 0.24/0.34 & N/A & \cite{stimulus-verification} \\
     Diverse Sampling & SGAN + NPSN  \cite{npsn} & CVPR 2022 & 0.50/0.96 & 13.03/23.04 &  \cite{npsn} \\
     Diverse Sampling & SGAN + BOSampler \cite{bosampler} & CVPR 2023 & 0.52/1.01 & N/A & \cite{bosampler} \\
     Diverse Sampling & SGAN + SV \cite{stimulus-verification} & CVPR 2023 & 0.49/0.94 & N/A & \cite{stimulus-verification} \\

    \bottomrule
    \end{tabular}
\end{table*}

\begin{table*}
    \caption{Comparisons of MTP Models on vehicle trajectory prediction benchmarks.}
    \label{tab:comparing-vehicle}
    \centering
    \begin{subtable}[h]{0.50\textwidth}
    \begin{tabular}{c|c||ccc} \toprule
     \multicolumn{5}{c}{Argoverse Leaderboard} \\ \midrule
     Framework & Method & minADE$_{6}$ & minFDE$_{6}$ & MR \\ \midrule \midrule
     Anchor (PEC) & HOME \cite{home} & 0.92 &  1.36 & 13.3\%\\
     Anchor (PEC) & TNT \cite{tnt} & 0.94 &  1.54 & 13.3\%\\
     Anchor (PEC) & DenseTNT \cite{densetnt} & 0.88 &  1.28 & 12.6\% \\
     Anchor (PTC) & MultiPath \cite{multipath} & 1.28 & N/A & N/A \\
    Anchor (PTC) & MultiPath++ \cite{multipath-pp} & 0.79 & 1.32 & 13.2\%\\
     Anchor (PTC) & QCNet \cite{qcnet} & 0.73 & 1.07 & 10.6\% \\
     
     \bottomrule
    \end{tabular}
    \end{subtable}
\hfill
    \begin{subtable}[h]{0.45\textwidth}
        \begin{tabular}{c|c||cc} \toprule
        \multicolumn{4}{c}{NuScenes Leaderboard} \\ \midrule
         Framework & Method & minFDE$_{5}$ & minFDE$_{10}$ \\ \midrule \midrule
         Noise (VAE) & AgentFormer \cite{agentformer}  & 1.86 & 1.45  \\
         Noise (VAE) & Trajectron++ \cite{trajectron-pp} & 1.88 & 1.51 \\
         Anchor (PEC) & GOHOME \cite{gohome} & 1.59 & 1.15 \\  
         Anchor (PEC) & SG-Net \cite{sgnet} & 1.86 & 1.40 \\
         Anchor (PTC) & MultiPath \cite{multipath} & 2.32 & 1.96 \\
         Anchor (PTC) & CoverNet \cite{covernet} & 1.96 & 1.48 \\
         \bottomrule
        \end{tabular}
    \end{subtable}
\end{table*}

\begin{table*}
\centering
\caption{Performance of MTP models on ForkingPath benchmark.}
\label{tab:comparing}

\begin{tabular}{c|c|ccccc}
    \toprule
     Frameworks & Model  & minADE$_{20\times N_{GT}}$ $\downarrow$ & minFDE$_{20\times N_{GT}}$  $\downarrow$ & Recall ($d=2cm$) $\uparrow$ & Precision ($d=2cm$) $\uparrow$ &  PTU $\uparrow$ \\ \midrule\midrule
     GAN (Noise) & SGAN \cite{social-gan}  & \textbf{1.33} & \textbf{2.12} & \textbf{30.00\%} & \textbf{50.68\%} & \textbf{55.65\%}  \\
     VAE (Noise) & AgentFormer \cite{agentformer}  &  1.64 & 2.95 & 25.87\% & 40.95\% & 53.76\%   \\
      PEC (PEC) & PECNet \cite{pecnet}  & 2.05 & 3.64 & 28.27\% & 29.45\% & 41.86\%  \\ 
     PEC (PEC) & YNet \cite{ynet}  &  1.73 & 2.52 & 22.17\% & 37.53\% & 53.08\% \\ 
     Diffusion (Noise) & MID \cite{mid}  & 1.89 & 3.45 & 24.25\% & 32.28\% & 48.25\%  \\
     PTC (PTC) & SIT \cite{social-interpretable-tree}  & 2.48 & 4.66 & 21.97\% & 19.86\% & 46.64\%  \\
     Recurrent (Grid) & Multiverse \cite{multiverse}  & 2.78 & 5.65 & 28.03\% & 13.74\% & 20.67\% \\ 
     \midrule

    PTC (PTC) & SIT \cite{social-interpretable-tree} (SDD) & 1.89 & 3.26 & 27.40\% & 31.67\% & 53.55\% \\
    
    PTC (PTC) & SIT \cite{social-interpretable-tree} (ETH)  & 2.26 & 4.16 & 21.50\%  & 21.06\% & 41.54\% \\

    \bottomrule     
\end{tabular}

\end{table*}
\subsection{Performance Analysis}
\noindent \textbf{Comparions on Commonly-used Datasets.} In pedestrian trajectory prediction, most studies have experimented with their models on ETH\&UCY and SDD datasets. In \Cref{tab:comparing-eth-ucy-sdd}, we report the minADE$_{20}$ and minFDE$_{20}$ performance of some typical methods for each MTP framework. All methods use the train/val/test splits in \cite{social-gan} for ETH\&UCY and splits in \cite{pecnet,ynet} for SDD.  At the current stage, STGlow \cite{stglow} performs the best among all MTP models without diverse sampling while the SGAN \cite{social-gan} performs the worst. YNet \cite{ynet} performs better than PECNet \cite{pecnet} on SDD due to the scene constraint for endpoint predictions. Finally, all diverse sampling methods provide an apparent performance boost, indicating that lower-bound based metrics can benefit from diverse predictions.

For vehicle trajectory prediction, we report in \Cref{tab:comparing-vehicle} the MTP model performance on Argoverse 1 Leaderboard and NuScenes benchmarks using the default metrics in their leaderboards. We find that studies in vehicle trajectory prediction favour anchor-based MTP framework. TNT \cite{tnt} has worse performance than \cite{home, tnt}, indicating that PEC models are sensitive to the sampling resolution. Then, the comparison between CoverNet and MultiPath illustrates that the quality of the prototype trajectory set largely influences the performance of PTC models. MultiPath++ \cite{multipath-pp}, QCNet \cite{qcnet} and ProphNet \cite{prophnet} all use generated prototypes outperform MultiPath, illustrating the advantage of learning-based prototype generation. 

\noindent \textbf{Comparions on ForkingPath Datasets.} Current benchmarks do not report distribution-aware metrics mainly because they provide only one future trajectory for each sample. Although ForkingPath annotates multi-future trajectories, to the best of our knowledge, ForkingPath is only experimented in a few studies \cite{multiverse,simaug,stmr,mggan-tp}. Therefore, we compare MTP methods the ForkingPath benchmark and report five metrics in \Cref{tab:comparing}, where min(A/F)DE$_{20\times N_{GT}}$ denotes that we compare 20 predictions with all annotated future trajectories for each sample and report the minimum error. We build the training set on \textit{ForkingPath-Anchor} \cite{multiverse, simaug}, a version fully reconstructing VIRAT \cite{virat} to CARLA environments with single-future annotation, and the testing set using the ForkingPath using the official splits in \cite{multiverse}. More details can be found in our {\color{blue} \href{https://github.com/HRHLALALA/Benchmark-on-ForkingPath}{code}}. 

Most MTP models have a maximum of 50.68\% of precision and 55.65\% of PTU, indicating that their predictions can cover only half of the annotated future trajectories. SGAN \cite{social-gan} achieves the best performance on all metrics. YNet \cite{ynet} performs better than PECNet \cite{pecnet}, indicating that scene interaction is important for PEC methods. SIT has worse performance than PECNet even though they share a similar experimental setup, which may because the prototype used in SIT reduces the model generalisation. However, we also find that SIT can be improved when being trained on the SDD and ETH training set, indicating that the dataset selection is essential for MTP. Multiverse \cite{multiverse} has the worst performance, mainly due to its low capability of predicting diverse trajectories.

However, ForkingPath only provides hundreds of testing cases and thus a better dataset is recommended to be explored. In addition, the performance of MTP methods is highly sensitive to the random seed, training strategies, training datasets and capability of feature extraction. Therefore, at the current stage, it is difficult to tell which MTP framework is the best. Further exploration is encouraged to analyse these MTP frameworks via controllable experiments.

 \section{Future Directions}
\label{sec:discussion} 
\subsection{Better Experimental Environment for MTP}
\noindent \textbf{Better Evaluation Metrics for MTP.} 
Evaluation metrics are important to assess the model architecture and loss functions. However, current metrics described in \cref{sec:evaluation-metrics} either fail to penalise unacceptable predictions or require huge efforts for annotations for other possible ground truths. Possible future work can look into evaluation metrics agnostic to ground truth. For example, a recent study \cite{ad-traj} suggests using abnormal detection models to help the decision-making of future trajectories which we believe can also be a future step for evaluation. Alternatively, rule-based metrics can be constructed to evaluate the social acceptance of all future trajectories. 

\noindent \textbf{Better Datasets for MTP.} 
As suggested in \cref{se:common-data}, MTP models require to be trained on large and representative datasets so that they can learn rich modalities of future trajectories. Alternatively, extra annotations for multiple modalities can be provided for MTP supervision. We suggest that future work can explore labour-efficient solutions to enhance the construction of the datasets for MTP. For example, metrics can be proposed to measure the richness of the modalities in a dataset and thereby helping the training set selection. In addition, we advocate for the exploration of more datasets like ForkingPath \cite{multiverse} to provide annotations with different modalities to enhance the evaluation of MTP. 
 
\noindent \textbf{Motion Planning using MTP.} We believe that MTP will eventually be used for downstream tasks such as motion planning and control in autonomous systems. The planning model can provide safe and collision-free routes based on trajectory predictions with multiple future predictions. However, to our knowledge, motion planning and MTP are currently developed independently. To connect these two fields, we first suggest exploring \textit{MTP-aware motion planning models} that can benefit from multi-future predictions. Then, the performance of these models can be used as an evaluation metric and \textit{planning-aware MTP models} can be investigated to help motion planning models achieve better results. 

\subsection{Robust, Efficient and Explainable MTP}
\noindent \textbf{Lightweight MTP Frameworks.} 
It is beneficial for MTP to predict trajectories with a longer horizon and a larger $K$ for more agents in parallel. Recently, some studies \cite{hypertraj, decouple-traj} tackle the efficiency issues in some models \cite{ynet, home} with the PEC framework, requiring large memory and latency consumption when predicting heatmaps for future trajectories using convolutional neural networks. Future works can further explore the lightweight MTP frameworks that can benefit real-time autonomous systems.

\noindent \textbf{Long-tail Problems.} Trajectory prediction datasets are distributed as long-tailed to some simple motions. \cite{lt-traj} shows that most cases are linear or even follow a constant velocity. Therefore, MTP models can easily overfit these easy cases and assign high probabilities to them. Although there are many models handling the long tail problem in trajectory prediction, \cite{eval-lt-traj} suggests that current models are too simple to provide a performance boost. \cite{bosampler} is the most recent sampling strategy that addresses this problem and we believe that further works can combine this sampling strategy with other solutions in computer vision for long tail problems. 

\noindent \textbf{MTP with Generalisation Problems.} 
MTP models may overfit the training set when the datasets are not large enough or not representative. Therefore, they can only generate predictions within limited and biased future trajectories. For example, if a model is trained on a scene when all paths are curved, it will provide multiple curved future trajectories even in an open space with no scene constraint. Some studies  \cite{t-gnn, causal-motion} tackles the domain adaptation and generalisation problem in trajectory prediction while some \cite{mp-traj, social-anchor} suggest that integrating human knowledge and deep learning models can largely enhance the robustness in unseen environments. In addition, SimAug \cite{simaug} explores the generalisation problem when using images with different views. Future directions can explore more advanced approaches for the OOD problems in MTP among scenes with different crowd densities \cite{t-gnn}, average speeds \cite{mp-traj}, terrains and views \cite{simaug}.

\noindent \textbf{Language Guided Explainable MTP.}
To build a trustable and safe autonomous system, it is essential to build explainable MTP frameworks that provide human-understandable decision-making for multi-future predictions. Currently, most MTP frameworks provide multiple predictions without explaining their decisions \cite{social-anchor}. Recently, Xue \textit{et.al.} \cite{prompt-time-series} have proposed a new prompt-based learning paradigm named PromptCast, suggesting that human language can be a prompt to guide time series forecasting with the form of questioning and answering. LMTraj \cite{lmtrajectory} also suggests that the language-based MTP model is powerful in dealing with social interactions. We believe that language can be used as prompts to guide multi-future predictions and provide human-readable explanations for predicted future trajectories. 

\subsection{Beyond MTP: Other Diverse Learning Tasks}
 Many other domains are facing the multi-solution problem: given the observed condition, there can be multiple plausible solutions. All these domains have similar goals with MTP, where the generated samples should be diverse, acceptable to humans and explainable/controllable. Therefore, techniques in these fields can be shared. 
    
 \noindent \textbf{Other Time-series Topics.} Some topics such as text, music or video-related problems also address diverse predictions. For example, in Neural Machine Translation, \cite{nas-nmt} finds that one sentence may have multiple plausible target translations, which causes the non-autoregressive model to alternatively provide words from two translated sentences. Also, diverse captioning tasks are proposed to solve multi-solution problems in image captions. \cite{grouptalk} suggests that different people can have different understandings of images and that generating appropriate, diverse, and high-quality descriptions of images is important for image captioning. \cite{metrics-diverse-captioning} proposes advanced evaluation metrics for the diversity of generated captions.


\noindent \textbf{Urban-wide MTP.} Current MTP methods focus on the future movement over small scales, e.g., in a street. However, we believe it can be extended to an urban-wide location prediction: \textit{human mobility prediction} (HMP), which is to predict the next place-of-interest (POI) based on previous locations and rich context information such as the semantics of locations. HMP is naturally multi-future since multiple POIs are acceptable with different uncertainties and hence the advances in MTP can be applied on it as well. Besides, MobTCast \cite{mobtcast} uses STP to enhance the geographical context for HMP. Therefore, we believe that MTP can be an even stronger auxiliary task to consider the uncertainty of future POIs.

\noindent \textbf{Human Behaviour Recognition.} This task is also called action or motion prediction. \HL{Unlike trajectory prediction that focuses on the human movements on the road only, it predicts human action labels and future actions given the past video frames \cite{human-activity-prediction, prediction-cgan} or skeletons \cite{lgn-action}.} To extract reliable latent features for future predictions, some methods \cite{lgn-action, prediction-cgan} apply advanced generative losses, although a single prediction is provided. A recent study DearNet \cite{dear-net} tackles uncertainty and diversity in this task. It estimates multiple action labels and conditionally generates multiple future skeletons. We note that some methods also predict action labels and multiple future trajectories \cite{action-contrastive-learning, titan}. Therefore, we believe that MTP can provide inspirations to future studies for this task.

\HL{
\noindent \textbf{MTP for Non-human Objects.} We suggest that MTP can also be used for motions of objects other than humans since there are complex factors that can change the trajectories. For example, Kapoor \textit{et al.} \cite{cyclone-tp} use variation RNN to model the uncertainty for cyclones, while C{\'\i}fka \textit{et al.} \cite{animal-tp} time-wisely model the uncertainty of the animal motions using Transformer, similar to the recurrent-based framework.}
 \section{Conclusion}
In this work, we provide the first comprehensive investigation for multi-future trajectory prediction, an important problem in trajectory prediction focusing on proposing multiple plausible future trajectories for each agent based on the past movement and the surrounding environment. Specifically, we build our taxonomies of existing frameworks, datasets and evaluation metrics for MTP and analyse in depth their benefits and challenges. We also compare the performance of typical MTP methods on commonly used benchmarks and report distribution-aware metrics on the ForkingPath dataset. Finally, we suggest several directions for future work such as better evaluation metrics and datasets, robust and lightweight MTP frameworks and sharing techniques in MTP to other problems.

\small{

    \bibliographystyle{IEEEtran}
    \bibliography{IEEEfull.bib}
}

\vfill
\end{document}